\documentclass[11pt]{article}

\usepackage[preprint]{acl}

\usepackage{amsmath,amsfonts,bm}

\def\eqref#1{equation~\ref{#1}}
\def\1{\bm{1}}

\DeclareMathAlphabet{\mathsfit}{\encodingdefault}{\sfdefault}{m}{sl}
\SetMathAlphabet{\mathsfit}{bold}{\encodingdefault}{\sfdefault}{bx}{n}

\usepackage{times}
\usepackage{latexsym}
\usepackage{paralist}
\usepackage{hyperref}
\usepackage{booktabs}
\usepackage{xspace}
\usepackage{pgfplots}

\usepackage[T1]{fontenc}
\usepackage[utf8]{inputenc}

\usepackage{microtype}

\usepackage{inconsolata}

\usepackage{hyperref}
\usepackage{url}
\usepackage{booktabs,tabularx}
\usepackage{graphicx}
\usepackage{booktabs}
\usepackage{multirow}
\usepackage{arydshln}
\usepackage[most]{tcolorbox}
\usepackage{xcolor}
\usepackage{enumitem}
\usepackage{subcaption}
\usepackage[font=small,labelfont=bf]{caption}
\pgfplotsset{compat=1.17}
\usepackage{tikz}
\usetikzlibrary{arrows.meta, positioning, shapes.geometric, fit, calc}

\definecolor{countryColor}{RGB}{0,153,51}   % Green
\definecolor{ruleColor}{RGB}{0,102,204}     % Blue
\definecolor{exampleColor}{RGB}{204,0,102}  % Dark Pink
\definecolor{ctxColor}{RGB}{0,102,204}       % blue
\definecolor{varColor}{RGB}{0,153,51}        % green
\definecolor{warnColor}{RGB}{204,0,102}      % magenta
\definecolor{boxBg}{RGB}{248,248,248}        % light gray
\tcbset{sharp corners, boxrule=0.5pt}

\usepackage{CJKutf8}
\usepackage{devanagari}

\title{Think Before You Link:\\Rarity, Reasoning, and Retrieval in Multilingual Entity Linking}

\author{
  Parinthapat Pengpun \quad Simran Khanuja \quad Graham Neubig \\
  \texttt{\{ppengpun, skhanuja, gneubig\}@andrew.cmu.edu} \\
  Carnegie Mellon University \\[0.4em]
  \url{https://neulab.github.io/think-before-you-link/}
}

\begin{document}
\maketitle

\newcommand{\refalg}[1]{Algorithm \ref{#1}}
\newcommand{\refeqn}[1]{Equation \ref{#1}}
\newcommand{\reffig}[1]{Figure \ref{#1}}
\newcommand{\reftbl}[1]{Table \ref{#1}}
\newcommand{\refsec}[1]{Section \ref{#1}}
\newcommand{\refapp}[1]{Appendix \ref{#1}}
\definecolor{coralpink}{rgb}{0.97, 0.51, 0.47}
\definecolor{babyblueeyes}{rgb}{0.63, 0.79, 0.95}

\newcommand{\jp}[1]{\begin{CJK}{UTF8}{min}#1\end{CJK}}
% Devanagari (Velthuis): argument is the devnag-preprocessed low-level form.
\newcommand{\dev}[1]{{\dn #1}}

\newcommand{\bmm}[1]{\bm{\mathcal{#1}}}
\newcommand{\real}[1]{\mathbb{R}^{#1}}
\newcommand{\method}{\textsc{CuE}\xspace}
\newcommand{\methodname}{\textsc{CuE}\xspace}
\newcommand{\mm}[2]{{\scriptsize (#1 / #2)}}

\newtheorem{theorem}{Theorem}[section]
\newtheorem{claim}[theorem]{Claim}

\newcommand\norm[1]{\left\lVert#1\right\rVert}

\newcommand{\note}[1]{\textcolor{blue}{#1}}

\newcommand*{\Scale}[2][4]{\scalebox{#1}{$#2$}}%
\newcommand*{\Resize}[2]{\resizebox{#1}{!}{$#2$}}%

\newcommand{\SK}[1]{\textcolor{cyan} {[\textsc{sk}: #1]}}

\newcommand{\tocite}{\textbf{\textcolor{blue} {[cite]}}}

% Pastel heatmap cells matching the visual palette of Figure 1.
% The endpoints are saturated enough to remain legible in print.
\definecolor{scorelow}{RGB}{244,188,132}
\definecolor{scoremid}{RGB}{252,249,245}
\definecolor{scorehigh}{RGB}{170,207,235}
\newcommand{\scorecell}[2]{%
  \begingroup
  \ifdim #1pt<50pt
    \pgfmathtruncatemacro{\scoremix}{100 - 2*(#1)}%
    \edef\scorecolor{scorelow!\scoremix!scoremid}%
  \else
    \pgfmathtruncatemacro{\scoremix}{2*(#1 - 50)}%
    \edef\scorecolor{scorehigh!\scoremix!scoremid}%
  \fi
  \setlength{\fboxsep}{1.2pt}%
  \colorbox{\scorecolor}{\strut #2}%
  \endgroup
}

% for teaser fig
\newcommand{\adjustimg}{% Horizontal adjustment of image
  \hspace*{\dimexpr\evensidemargin-\oddsidemargin}%
}
\newcommand{\centerimg}[2][width=\textwidth]{% Center an image
  \makebox[\textwidth]{\adjustimg\includegraphics[#1]{#2}}%
}

\begin{abstract}
Multimodal entity linking grounds entity mentions in text and images to knowledge-base entries. These systems degrade on rare entities, but prior work measures rarity primarily through popularity-based metrics such as pageviews. We broaden this view using knowledge-graph structural metrics that capture how well an entity is documented and connected. These metrics identify many rare entities that popularity metrics miss. Across the resulting rare-entity slices, state-of-the-art accuracy drops by 15.4--39.9\%, showing that different rarity definitions expose different failure modes. To address these failures, we introduce a simple, training-free framework in which a reasoning-capable vision-language model iteratively searches and reasons over Wikipedia, gathering evidence dynamically. Controlled experiments show that reasoning and retrieval are complementary. Reasoning alone does not significantly improve accuracy on rare entities. Retrieval without reasoning improves rare-entity accuracy but can hurt overall accuracy. Their combination performs best. On MERLIN, a multilingual multimodal entity linking benchmark over five languages (Hindi, Indonesian, Japanese, Tamil, Vietnamese), our best system improves over the state of the art by 6.9\% overall and by up to 23.3\% on rare-entity slices. We release MERLIN-Rare, rare-entity test slices for targeted evaluation, with our framework.
\end{abstract}

\section{Introduction}

\begin{figure}[t!]
\centering
\includegraphics[width=\columnwidth]{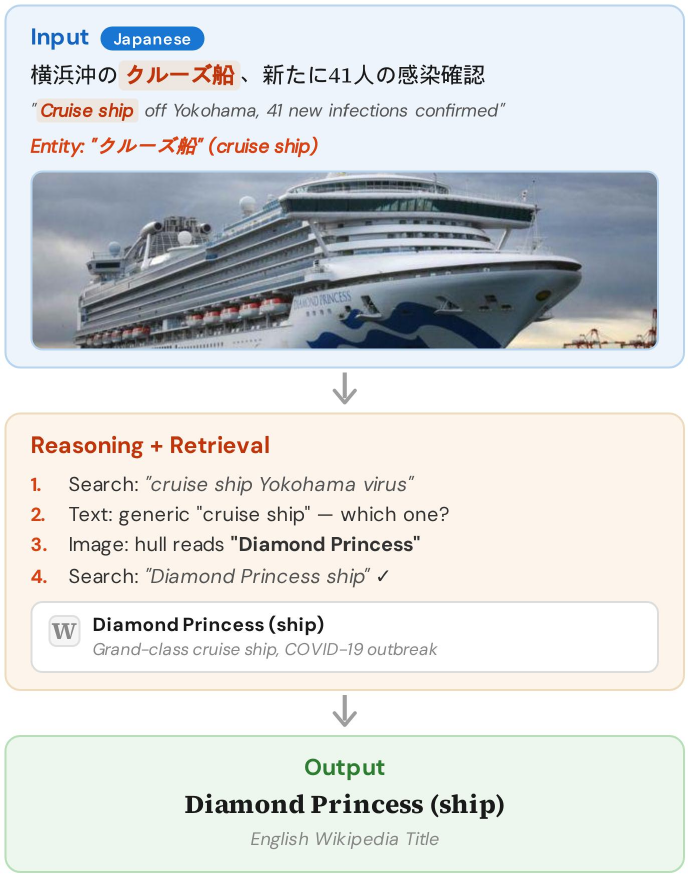}
\caption{Multimodal entity linking. The Japanese text mentions a generic ``cruise ship,'' which cannot be resolved from text alone. The accompanying image shows the vessel with its name ``Diamond Princess'' written on the hull, enabling a targeted search to link to the correct entity.}
\label{fig:task-example}
\end{figure}

Entity linking, or the task of grounding textual mentions to knowledge base entries, is
foundational for knowledge-intensive NLP applications \cite{Sevgili_2022, 6823700}.
As vision-language models become increasingly capable, there is growing interest in
multimodal entity linking, where visual context can help disambiguate
mentions that would be ambiguous from text alone \cite{shi-etal-2024-generative, moon-etal-2018-multimodal-named}.
This is particularly valuable in multilingual settings, where images provide
language-agnostic signal that can bridge gaps in textual coverage \cite{ramamoorthy2025merlintestbedmultilingualmultimodal}.

\begin{figure*}[t!]
% \vspace{-1em}
\centering
\includegraphics[width=0.85\textwidth]{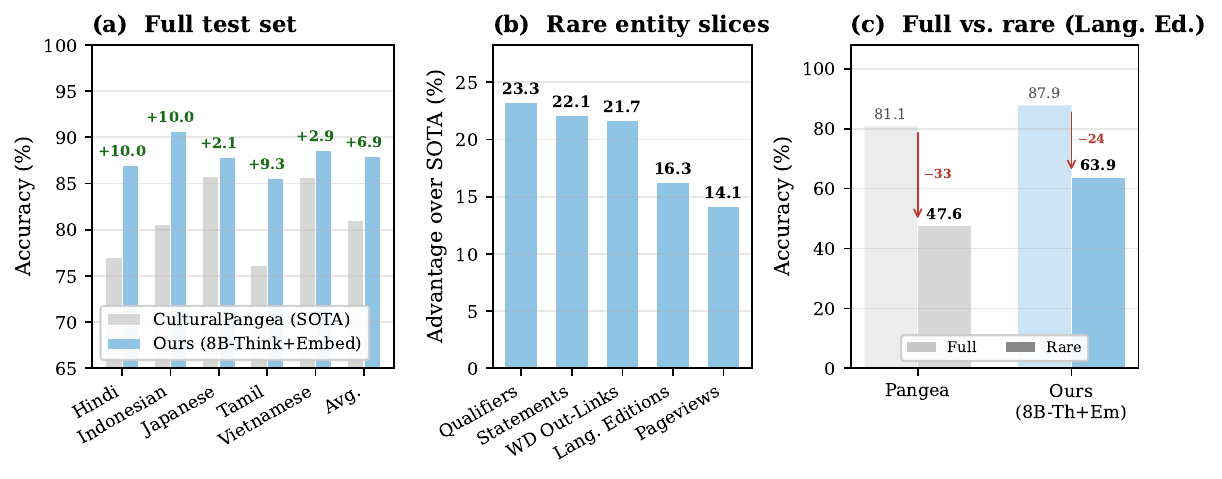}
\vspace{-0.5em}
\caption{\textbf{Our best system (8B-Think+Embed) vs.\ baselines.} \textbf{(a)} Full test set accuracy across five languages: +6.9\% average over SOTA. \textbf{(b)} Advantage over Pangea on bottom-5\% rare entity slices reaches \textbf{+23.3\%}. \textbf{(c)} On rare entities in the bottom 5\% by language editions, Pangea reaches 47.6\% while our system reaches 63.9\%.}
\label{fig:hero}
% \vspace{-1em}
\end{figure*}

Currently, entity linking models perform
well on common, frequently-referenced entities but degrade sharply on the
long tail \cite{boscariol2025evaluationllmslongtailentity, hoveyda-etal-2024-real, ilievski-etal-2018-systematic}.
This problem is especially severe for culturally niche entities, or those well-known within specific language communities but sparsely represented in cross-lingual knowledge bases~\cite{veselovsky2025localizedculturalknowledgeconserved, naous2024havingbeerprayermeasuring}, as opposed to entities that are merely unpopular. Prior work, however, has gauged entity rarity through popularity-based proxies such as Wikipedia pageviews and incoming link counts~\citep{Graciotti_2025, mallen-etal-2023-trust, chen-etal-2021-evaluating}, which capture access frequency but may not reflect cultural specificity.

We study this problem on MERLIN~\cite{ramamoorthy2025merlintestbedmultilingualmultimodal}, a multilingual multimodal entity linking benchmark spanning five languages (Hindi, Indonesian, Japanese, Tamil, and Vietnamese). We propose a broader characterization of entity rarity that distinguishes cultural specificity from mere unpopularity. Using knowledge-graph structural metrics, we show that structural sparsity is associated with substantial degradation on entities that popularity-based metrics often miss. The bottom-5\% entity sets overlap by only 37\% on average, and accuracy declines toward the structurally sparse end of the distribution.

Intuitively, different rarity definitions reveal different failure modes. An entity may receive little traffic despite rich documentation, or it may be popular within one community but have sparse cross-lingual and graph coverage. A single popularity metric cannot distinguish these cases. Hence, we consider multiple definitions of rarity to identify failure modes hidden by aggregate evaluation.

On the current state-of-the-art~\cite{nyandwi2025groundingmultilingualmultimodalllms} model, accuracy drops by 15.4--39.9\% across the bottom-5\% slices. Structural metrics reveal drops of up to 37.0\%, similar to 37.7\% for pageviews, but identify largely different entities. Thus, standard popularity metrics miss many rare entities on which the model fails.

Rare entities are unlikely to be well-represented in model weights, motivating retrieval-augmented approaches that access external knowledge at inference time \cite{ding2025entgptentitylinkinggenerative, Pons_2024, liu2024onenetfinetuningfreeframework}. However, retrieval alone may be insufficient as entity disambiguation often involves contextual reasoning. We investigate whether a simple framework combining reasoning-capable vision-language models with retrieval over Wikipedia can address rare entity failures.

Using this framework, we find that reasoning and retrieval play complementary roles. Reasoning alone does not significantly improve accuracy on rare entities. Retrieval improves performance on rare entities even without reasoning, but it hurts the non-reasoning model on the full test set. Their combination gives the strongest performance, which suggests that reasoning helps the model use retrieved evidence more effectively.

As shown in Figure~\ref{fig:hero}, our best system achieves improvements of \textbf{+2.1\%} to \textbf{+10.0\%} over the SOTA baseline across languages.
The gains are larger on rare entities with the full-dataset advantage of \textbf{+6.9\%} growing to \textbf{+23.3\%} on the hardest rare entity slices.
In summary, our contributions are:
\begin{compactitem}
    \item \textbf{Rarity characterization:} We propose a multidimensional 
    characterization of entity rarity using Wikidata structural metrics, 
    showing that different rarity definitions identify distinct entity sets
    and failure modes. We release
    \textsc{Merlin-Rare}, a set of rare entity test slices for targeted 
    evaluation.
    \item \textbf{Framework:} We present a simple framework combining 
    reasoning-capable VLMs with iterative retrieval over Wikipedia, achieving 
    \textbf{+6.9\%} over the state-of-the-art on MERLIN and
    up to \textbf{+23.3\%} on the rare entity slices.
    \item \textbf{Analysis:} We provide a detailed analysis of model behavior showing that, retrieval  hurts non-reasoning models on common entities but helps on rare 
    ones. We also show that  retrieval failure accounts for 72\% of residual errors, and that 
    reasoning models make fewer but more targeted retrieval calls.
\end{compactitem}

% \noindent We release our code and evaluation scripts to support reproducibility.

\section{Related Work}

\paragraph{Entity Linking with Large Language Models.}
Entity linking has shifted from classification over candidate sets to direct generation, beginning with GENRE's~\cite{decao2021autoregressiveentityretrieval} autoregressive entity retrieval and extended via context enrichment and adaptive routing~\cite{ding-etal-2024-chatel, ding2025entgptentitylinkinggenerative, xin2025llmaellargelanguagemodels, liu2024onenetfinetuningfreeframework, li-etal-2025-leveraging-power}. LELA~\cite{haffoudhi2026lelallmbasedentitylinking} retrieves once then reasons with self-consistency voting; ELA~\cite{luo2025entitylinkingagentquestion} uses a single retrieval call with no query refinement. All operate text-only and predominantly in English; none investigate iterative evidence gathering for culturally niche entities.

\paragraph{Multimodal and Multilingual Entity Linking.}
Multimodal EL leverages visual context to disambiguate text-ambiguous mentions~\cite{moon-etal-2018-multimodal-named, wang-etal-2022-wikidiverse, shi-etal-2024-generative, Liu_2025}, while multilingual EL has advanced via autoregressive and end-to-end methods~\cite{decao2021multilingualautoregressiveentitylinking, limkonchotiwat-etal-2023-mrefined}. MERLIN~\cite{ramamoorthy2025merlintestbedmultilingualmultimodal} is one of the first benchmarks at this intersection; Cultural Pangea~\cite{nyandwi2025groundingmultilingualmultimodalllms} establishes the SOTA by fine-tuning a multilingual VLM on culturally grounded data, yet still degrades sharply on structurally rare entities (Section~\ref{sec:rarity}).

\paragraph{Entity Rarity and Cultural Representation.}
EL systems degrade on rare entities~\cite{ilievski-etal-2018-systematic, hoveyda-etal-2024-real, boscariol2025evaluationllmslongtailentity}, with rarity typically equated with low popularity~\cite{mallen-etal-2023-trust,kandpal2023largelanguagemodelsstruggle, chen-etal-2021-evaluating}. However, unpopularity is not the same as cultural specificity. LLMs exhibit Western-centric entity bias~\cite{naous2024havingbeerprayermeasuring}, VLM performance correlates with per-language Wikipedia size~\cite{pmlr-v162-bugliarello22a}, and localized cultural knowledge is poorly represented cross-lingually~\cite{veselovsky2025localizedculturalknowledgeconserved, tao2024cultural, adilazuarda2024towards}. We distinguish these through a multidimensional rarity characterization separating structural sparsity from popularity.

\paragraph{Retrieval-Augmented Reasoning.}
ReAct~\cite{yao2023reactsynergizingreasoningacting} and IRCoT~\cite{trivedi2023interleavingretrievalchainofthoughtreasoning} interleave reasoning with retrieval; reasoning-native models trained via RL~\cite{Guo_2025, qwen3technicalreport, jin2025searchr1trainingllmsreason, feng2025retoolreinforcementlearningstrategic} learn to call search within their thinking, and inference-time compute can substitute for parameters~\cite{snell2024scalingllmtesttimecompute}. This paradigm has not been applied to entity linking, which is the gap we close.

\section{Task Definition}
\label{sec:task_def}

Entity linking is the task of mapping textual entity mentions to entries in a knowledge base~\cite{6823700}. We study a multilingual, multimodal formulation of this task, focusing on entity linking given a marked mention. We utilize MERLIN~\cite{ramamoorthy2025merlintestbedmultilingualmultimodal} as our evaluation set.

We follow MERLIN's setup.
Given a text passage $T$ in a source language, an accompanying image $I$, 
and a marked entity mention $m \in T$, the task is to predict the 
English Wikipedia title of the entity referenced by $m$. We evaluate 
using exact-match accuracy against gold annotations.

% \textbf{Input: }The input consists of three components: (1) a text passage $T$ in a source language, (2) a single image $I$ accompanying the text, and (3) an entity mention $m$ marked within $T$. The text and image together provide context for disambiguation.

% \textbf{Output: }The output is the English Wikipedia title of the entity referenced by $m$. While source texts may appear in various languages, the target knowledge base is English Wikipedia.

% \textbf{Evaluation Metric: } We evaluate using exact match accuracy against gold annotations.
\section{Entity Rarity Analysis}
\label{sec:rarity}

Before presenting our methodology, we characterize the rarity problem that motivates our approach. We define entity rarity along multiple dimensions and show that the current state-of-the-art fails on rare entities.

\subsection{Defining Entity Rarity}
\label{subsec:rarity}

Prior work commonly defines entity rarity using popularity signals such as Wikipedia pageviews or incoming link counts~\cite{Graciotti_2025, xin2025llmaellargelanguagemodels, mallen-etal-2023-trust, ilievski-etal-2018-systematic}. However, popularity is only one dimension of rarity. Intuitively, an entity can receive substantial public attention but still have limited structured or cross-lingual information. Wikipedia content metrics measure how much an entity has been documented, while Wikidata metrics measure its structural connectivity and coverage across languages. These dimensions can reflect different sources of difficulty for entity linking. Limited documentation reduces the available textual evidence, sparse knowledge-graph structure provides fewer relations between entities, and low cross-lingual coverage makes it harder to connect a source-language mention to an English knowledge-base entry. Prior work also shows that Wikipedia attention can differ from Wikidata structure~\citep{erenrich2024psych}, and that culturally contextual content often has limited coverage across language editions~\citep{miquelribe2018wikipedia}. Popularity-based definitions may miss entities that receive attention but remain poorly represented in the resources used by multilingual EL systems. We consider multiple definitions of rarity to identify model failure modes that popularity-based metrics may not reveal.

\paragraph{Rarity Metrics.}
We collect two families of metrics via the Wikipedia and Wikidata APIs.
\textbf{Wikipedia-based metrics} reflect editorial attention and documentation depth: pageviews (90-day), backlinks, article size, revision count, unique editors, category count, external links, reference count, and image count.
\textbf{Wikidata-based metrics} reflect structural connectivity and cross-lingual coverage: incoming links, outgoing links, language editions (number of Wikipedia languages with an article), statement count, qualifier count, and entity age.

\paragraph{Definitions.}
An entity $e$ is \emph{rare on metric} $m$ if $m(e)$ falls in the bottom $q\%$ of the test-set distribution ($q{=}5$ in the main text). An entity is \emph{unpopular} if it is rare on access-frequency metrics (pageviews, backlinks), and \emph{structurally rare} if it is rare on Wikidata metrics (language editions, statements, qualifiers, links).

We use rare as an umbrella for these metric-specific tails, which include unpopular entities with low access frequency, under-documented entities with limited Wikipedia content, and structurally sparse entities with limited Wikidata coverage. We use the bottom 5\% in the main analysis because it balances rarity severity with enough examples for reliable evaluation. Appendix~\ref{app:threshold-robustness} shows that the findings remain stable at 1\%, 5\%, and 10\%, and Appendix~\ref{app:deciles} shows the same gradient across rarity deciles.

These dimensions are complementary. An entity could be popular yet structurally rare. For example, the entity \emph{2016 Indian banknote demonetisation} is editorially rich but structurally sparse, while the \emph{Muttahida Qaumi Movement} is the reverse, a thin English article atop a dense knowledge graph. Empirically, the bottom-5\% entity sets for different metrics share only 37\% of their entities on average (Appendix~\ref{app:metric-overlap}), with some pairs overlapping as little as 10\%.

We further note that cross-lingual knowledge base coverage tracks cultural representation in models. VLM accuracy on multilingual benchmarks scales with per-language Wikipedia size~\citep{pmlr-v162-bugliarello22a}, LLMs default to English-centric outputs even when prompted in other languages~\citep{veselovsky2025localizedculturalknowledgeconserved, naous2024havingbeerprayermeasuring}, and digitally underrepresented cultures receive ``thin descriptions'' that amplify downstream bias~\citep{adilazuarda2024towards}. We therefore treat structural sparsity in cross-lingual signals as a culturally meaningful rarity signal.

\subsection{Baseline Degradation on Rare Entities}
\label{subsec:baseline-rarity}

We evaluate Cultural Pangea~\citep{nyandwi2025groundingmultilingualmultimodalllms}, the current state-of-the-art on MERLIN (81.1\% avg), on bottom-5\% entity slices for each metric. Crucially, Cultural Pangea was explicitly trained on culturally-grounded data, making it a strong test case for examining whether rarity remains problematic for models designed to handle diverse entities. For each rarity metric, we compute accuracy for both systems on the same bottom-5\% entity set.

\begin{table}[b!]
\centering
\footnotesize
\setlength{\tabcolsep}{3.5pt}
\begin{tabular}{lrrrrr}
\toprule
\textbf{Evaluation Slice} & \textbf{Hi} & \textbf{Id} & \textbf{Ja} & \textbf{Ta} & \textbf{Vi} \\
\midrule
Full Test Set & \scorecell{77.0}{77.0} & \scorecell{80.6}{80.6} & \scorecell{85.8}{85.8} & \scorecell{76.2}{76.2} & \scorecell{85.7}{85.7} \\
\midrule
\multicolumn{6}{l}{\textit{Wikipedia-based metrics (Bottom 5\%)}} \\
Pageviews (90d) & \scorecell{42.0}{42.0} & \scorecell{52.9}{52.9} & \scorecell{41.9}{41.9} & \scorecell{30.6}{30.6} & \scorecell{49.2}{49.2} \\
Backlinks & \scorecell{56.7}{56.7} & \scorecell{55.9}{55.9} & \scorecell{39.5}{39.5} & \scorecell{43.5}{43.5} & \scorecell{54.0}{54.0} \\
Article Size & \scorecell{34.8}{34.8} & \scorecell{62.9}{62.9} & \scorecell{48.8}{48.8} & \scorecell{40.3}{40.3} & \scorecell{48.4}{48.4} \\
Revision Count & \scorecell{34.8}{34.8} & \scorecell{55.7}{55.7} & \scorecell{44.2}{44.2} & \scorecell{35.5}{35.5} & \scorecell{53.1}{53.1} \\
Unique Editors & \scorecell{37.7}{37.7} & \scorecell{55.7}{55.7} & \scorecell{42.4}{42.4} & \scorecell{35.5}{35.5} & \scorecell{51.6}{51.6} \\
Category Count & \scorecell{24.1}{24.1} & \scorecell{55.4}{55.4} & \scorecell{41.1}{41.1} & \scorecell{40.0}{40.0} & \scorecell{45.3}{45.3} \\
External Links & \scorecell{32.8}{32.8} & \scorecell{60.0}{60.0} & \scorecell{41.9}{41.9} & \scorecell{33.9}{33.9} & \scorecell{50.8}{50.8} \\
Reference Count & \scorecell{32.8}{32.8} & \scorecell{59.7}{59.7} & \scorecell{47.6}{47.6} & \scorecell{38.6}{38.6} & \scorecell{56.3}{56.3} \\
Image Count & \scorecell{53.4}{53.4} & \scorecell{62.7}{62.7} & \scorecell{52.6}{52.6} & \scorecell{43.6}{43.6} & \scorecell{56.7}{56.7} \\
\midrule
\multicolumn{6}{l}{\textit{Wikidata-based metrics (Bottom 5\%)}} \\
Incoming Links & \scorecell{62.1}{62.1} & \scorecell{65.2}{65.2} & \scorecell{51.2}{51.2} & \scorecell{52.6}{52.6} & \scorecell{56.5}{56.5} \\
Outgoing Links & \scorecell{45.2}{45.2} & \scorecell{44.1}{44.1} & \scorecell{47.4}{47.4} & \scorecell{43.5}{43.5} & \scorecell{42.6}{42.6} \\
Language Editions & \scorecell{46.4}{46.4} & \scorecell{50.0}{50.0} & \scorecell{44.2}{44.2} & \scorecell{48.3}{48.3} & \scorecell{49.2}{49.2} \\
Statement Count & \scorecell{41.8}{41.8} & \scorecell{46.3}{46.3} & \scorecell{47.6}{47.6} & \scorecell{41.7}{41.7} & \scorecell{42.9}{42.9} \\
Qualifier Count & \scorecell{44.1}{44.1} & \scorecell{47.6}{47.6} & \scorecell{50.0}{50.0} & \scorecell{39.5}{39.5} & \scorecell{44.4}{44.4} \\
Entity Age & \scorecell{72.5}{72.5} & \scorecell{77.3}{77.3} & \scorecell{69.8}{69.8} & \scorecell{40.3}{40.3} & \scorecell{68.4}{68.4} \\
\bottomrule
\end{tabular}
\caption{Cultural Pangea accuracy (\%) on the full MERLIN test set and bottom-5\% slices for each rarity metric. Pastel orange cells indicate lower accuracy, while pastel blue cells indicate higher accuracy.}
\label{tab:baseline-rarity}
% \vspace{-1.5em}
\end{table}

\begin{figure}[t]
\centering
\includegraphics[width=\columnwidth]{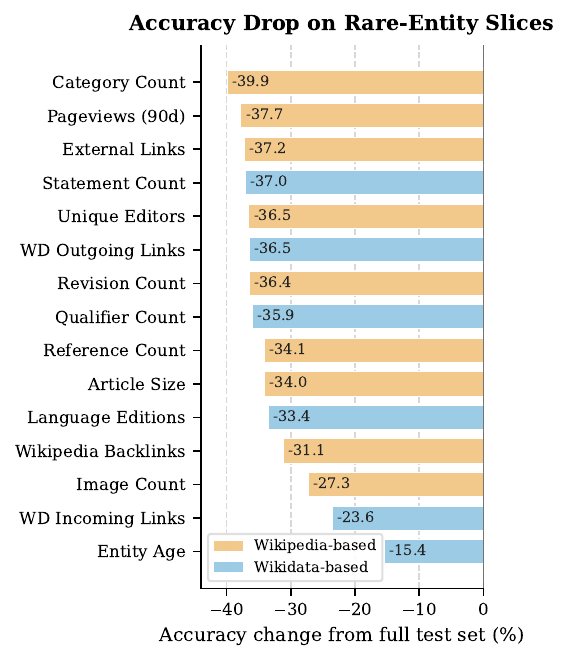}
\vspace{-1.5em}
\caption{Cultural Pangea accuracy change (\%) on bottom-5\% slices relative to the full test set. Negative values indicate degradation.}
\label{fig:metric-drops}
\end{figure}

Figure~\ref{fig:metric-drops} summarizes CulturalPangea's degradation across all rare-entity subsets. Relative to its 81.1\% full-set accuracy, its drop ranges from 15.4\% to 39.9\% across the bottom-5\% subsets. This degradation is not confined to popularity-based tails. Accuracy drops by 37.7\% on the pageview slice and by 37.0\% on the Wikidata statement-count slice. Since the metric-defined tails overlap by only 37\% on average (Appendix~\ref{app:metric-overlap}), popularity-only evaluation would miss many structurally sparse entities on which the baseline suffers comparable degradation. Table~\ref{tab:baseline-rarity} provides a per language breakdown of this degradation. GEMEL and mGENRE show the same overall pattern in Appendix~\ref{app:baseline-rarity}.

% To facilitate targeted evaluation on rare entities, we will release \textbf{MERLIN-Rare}, bottom-5\% entities for each metric across all five languages, upon acceptance.

\section{Methodology}
\label{sec:method}
Culturally niche entities are those least likely to be well-represented in model parameters, since training data skews toward well-documented entities. Furthermore, disambiguating rare entities may require multi-step reasoning~\cite{trivedi2023interleavingretrievalchainofthoughtreasoning}. Hence, we propose a simple framework with a reasoning-capable VLM with iterative retrieval over external knowledge sources. 

%The model performs multi-step reasoning, issuing search queries and incorporating retrieved evidence until confident. We evaluate both lexical and semantic retrieval to study how retrieval quality affects entity linking, particularly for rare entities.
% Crucially, our architecture is model-agnostic: the same pipeline can evaluate both reasoning-native models and instruction-tuned models, allowing us to isolate the contribution of each capability.

\subsection{Reasoning with Retrieval}

\paragraph{Model Selection.}
% We use the Qwen3-VL model family~\cite{qwen3technicalreport}, one of the best-performing open-source VLMs, in both Thinking (reasoning-native) and Instruct variants at 2B, 4B, and 8B parameter sizes. This choice is motivated by three requirements: (1) strong multilingual understanding across our target languages, (2) explicit reasoning support for multi-step disambiguation in the Thinking variant, and (3) vision-language capability to process news article images alongside text. 
%For inference, we use the SGLang~\cite{zheng2024sglangefficientexecutionstructured} engine to support batched inference.
We use the Qwen3-VL model family~\cite{qwen3technicalreport}, one of the best-performing open-source VLMs, in both Thinking (reasoning-native) and Instruct variants at 2B, 4B, and 8B parameter sizes. This family uniquely provides matched architecture across reasoning and non-reasoning variants at multiple scales, enabling controlled comparisons that isolate the contributions of reasoning, retrieval, and model size. 
%We additionally require strong multilingual understanding across our target languages and robust vision-language capability.

\paragraph{Retrieval System.}
While rare entities are unlikely to be encoded in model parameters, they may still be documented in external knowledge sources. Retrieval-augmented approaches can bridge this gap by accessing such sources at inference time. We use English Wikipedia as our retrieval corpus and evaluate two retrieval strategies.
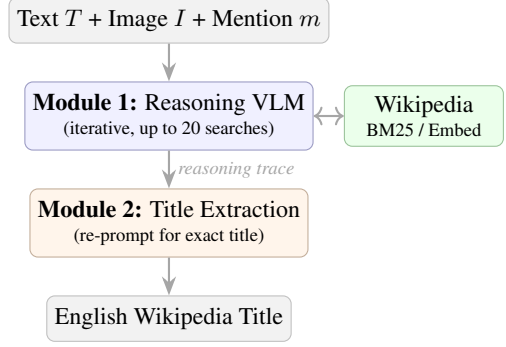
\begin{figure}[t]
\centering
\small
\begin{tikzpicture}[
    node distance=0.5cm,
    box/.style={draw, rounded corners=3pt, minimum width=3.2cm, minimum height=0.6cm, align=center, font=\footnotesize},
    io/.style={box, fill=gray!10, draw=gray!50},
    Module/.style={box, minimum height=0.9cm},
    retriever/.style={box, fill=green!8, draw=green!40!black!40, minimum width=2.1cm, minimum height=0.8cm},
    arr/.style={-{Stealth[length=2.5mm]}, thick, color=gray!70},
    label/.style={font=\scriptsize\itshape, color=gray!70},
]
% Input
\node[io] (input) {Text $T$ + Image $I$ + Mention $m$};

% Module 1
\node[Module, fill=blue!6, draw=blue!40!black!40, below=of input] (l1) {%
    \textbf{Module 1:} Reasoning VLM\\[-1pt]
    {\scriptsize (iterative, up to 20 searches)}
};

% Retriever
\node[retriever, right=0.4cm of l1] (ret) {%
    Wikipedia\\[-1pt]
    {\scriptsize BM25 / Embed}
};

% Module 2
\node[Module, fill=orange!8, draw=orange!40!black!40, below=of l1] (l2) {%
    \textbf{Module 2:} Title Extraction\\[-1pt]
    {\scriptsize (re-prompt for exact title)}
};

% Output
\node[io, below=of l2] (output) {English Wikipedia Title};

% Arrows
\draw[arr] (input) -- (l1);
\draw[arr] (l1) -- node[right, label] {reasoning trace} (l2);
\draw[arr] (l2) -- (output);

% Bidirectional arrow between Module 1 and Retriever
\draw[arr, <->] (l1.east) -- (ret.west);

\end{tikzpicture}
\caption{Two-module pipeline. Module 1 performs iterative reasoning with retrieval access over Wikipedia. Module 2 re-prompts the model to extract the final title from the reasoning.}
\label{fig:pipeline}
% \vspace{-1em}
\end{figure}

\textbf{BM25 (Lexical).} We built a retrieval system using the
\texttt{wikimedia/structured-wikipedia} dataset from Hugging
Face,
%\footnote{\href{https://huggingface.co/datasets/wikimedia/structured-wikipedia}{Dataset URL}}
which provides pre-processed Wikipedia dumps for the English
language. We indexed articles using BM25~\citep{bm25s}, a
lexical retrieval method that matches query terms against document terms. BM25 is fast and effective when query terms overlap with target titles, however it has a clear flaw when entity mentions appear in non-Latin scripts and must be transliterated to match English Wikipedia titles for searching.

\textbf{Embedding (Semantic).} To address the cross-lingual limitation of BM25, we also evaluate semantic retrieval using \texttt{intfloat/multilingual-e5-large-instruct} \cite{wang2024multilinguale5}, a multilingual embedding model. We embed each English Wikipedia title-description pair into a FAISS index~\cite{douze2024faiss}. The search string is the query, and the top-$k$ nearest pairs are returned as snippets.

\begin{table*}[t!]
\centering
\small
\begin{tabular}{llrrrrrr}
\toprule
\textbf{Model} & \textbf{RAG} & \textbf{Hindi} & \textbf{Indo.} & \textbf{Japan.} & \textbf{Tamil} & \textbf{Viet.} & \textbf{Avg.} \\
\midrule
\multicolumn{8}{l}{\textit{Prior work}} \\
GEMEL & --- & 55.5 & 71.6 & 73.2 & 23.9 & 69.4 & 58.7 \\
mGENRE & --- & 59.4 & 84.3 & 76.7 & 68.1 & 75.7 & 72.9 \\
CulturalPangea-7B & --- & 77.0 & 80.6 & 85.8 & 76.2 & 85.7 & 81.1 \\
\midrule
\multicolumn{8}{l}{\textit{Retrieval-aware baseline}} \\
CulturalPangea-RAG & Embed & 67.7 & 80.0 & 86.3 & 55.4 & 83.1 & 74.5 \\
\midrule
\multicolumn{8}{l}{\textit{Ours}} \\
2B-Instr & --- & 62.2 & 74.7 & 64.4 & 50.4 & 65.8 & 63.5 \\
2B-Instr & BM25   & 47.7 & 63.3 & 54.4 & 41.7 & 60.9 & 53.6 \\
2B-Instr & Embed  & 54.8 & 62.1 & 55.6 & 43.1 & 55.7 & 54.3 \\
2B-Think & --- & 62.5 & 74.2 & 63.6 & 45.6 & 67.7 & 62.7 \\
2B-Think & BM25   & 62.8 & 68.2 & 62.1 & 45.7 & 63.3 & 60.4 \\
2B-Think & Embed  & 59.1 & 69.3 & 59.3 & 46.0 & 61.6 & 59.1 \\
\midrule
4B-Instr & --- & 75.4 & 79.7 & 75.6 & 71.2 & 78.9 & 76.2 \\
4B-Instr & BM25   & 74.1 & 79.0 & 67.6 & 70.6 & 72.3 & 72.7 \\
4B-Instr & Embed  & 76.7 & 83.8 & 70.7 & 73.4 & 76.1 & 76.1 \\
4B-Think & --- & 77.9 & 83.5 & 79.8 & 72.7 & 84.2 & 79.6 \\
4B-Think & BM25   & 78.9 & 85.6 & 80.5 & 73.9 & 84.6 & 80.7 \\
4B-Think & Embed  & 83.0 & 89.6 & 82.9 & 76.4 & 86.7 & 83.7 \\
\midrule
8B-Instr & --- & 81.4 & 86.6 & 83.0 & 80.6 & 85.7 & 83.5 \\
8B-Instr & BM25   & 77.4 & 80.8 & 79.4 & 75.4 & 79.8 & 78.6 \\
8B-Instr & Embed  & 80.0 & 86.4 & 86.1 & 77.3 & 84.9 & 82.9 \\
8B-Think & --- & 81.7 & 86.8 & 84.2 & 82.0 & 86.1 & 84.2 \\
8B-Think & BM25   & 82.8 & 88.0 & 86.6 & 84.4 & \underline{87.4} & 85.8 \\
8B-Think & Embed  & \textbf{87.0} & \textbf{90.6} & \textbf{87.9} & \textbf{85.5} & \textbf{88.6} & \textbf{87.9} \\
\midrule
\multicolumn{2}{l}{$\Delta$ \textit{best vs.\ Pangea}} & \textit{+10.0} & \textit{+10.0} & \textit{+2.1} & \textit{+9.3} & \textit{+2.9} & \textit{+6.9} \\
\bottomrule
\end{tabular}
\caption{Accuracy (\%) on the full MERLIN test set. \textbf{Bold} = best per column, \underline{underline} = second best. Our best system (8B-Think+Embed) outperforms Cultural Pangea by +6.9\% on average. Reasoning models consistently outperform instruct models. Embedding retrieval outperforms BM25. BM25 hurts instruct models but helps reasoning models.}
% \vspace{-0.9em}
\label{tab:main-results}
\end{table*}

\paragraph{Implementation.}
The pipeline is summarized in Figure~\ref{fig:pipeline}. Given an input image $I$, text passage $T$, and entity mention $m$, the model performs iterative reasoning with retrieval access. At each step, the model will:
\begin{compactenum}
    \item Analyze the visual and textual context to identify disambiguating signals
    \item Issue a search query to the retrieval system
    \item Incorporate retrieved Wikipedia snippets into its reasoning
    \item Repeat until confident in a final answer
\end{compactenum}

\noindent We force the first search call in every RAG configuration. In preliminary runs, the 2B models often produced an answer without calling the tool despite explicit instructions to search. Without this control, a nominal RAG configuration could behave like No RAG. After the first call, tool use is automatic, and the model decides whether and how to continue searching. The model is allowed up to 20 retrieval iterations per example. It produces a freeform explanation of its reasoning process, including which candidate entities it considered and why it selected the final answer. A second pass then re-prompts the model with its own complete reasoning trace and instructs it to output only the final Wikipedia title.

% Figure~\ref{fig:pipeline} summarizes the two-module pipeline.

% We note that a preliminary single-pass variant where the model was instructed to only emit its final answer at the end of the reasoning trace produced verbose or hedged outputs (restating multiple candidates, appending qualifiers) that failed exact match despite containing the correct entity.

\section{Experimental Setup}
\label{sec:setup}

\paragraph{Model variants.} We evaluate Qwen3-VL~\cite{qwen3technicalreport} in two variants: \textbf{Thinking} (reasoning-native, trained with reinforcement learning to produce extended reasoning traces) and \textbf{Instruct} (standard instruction-tuned). Both share the same base architecture and are evaluated at 2B, 4B, and 8B parameter sizes. 

\paragraph{Retrieval methods.} Each model variant is evaluated under three retrieval conditions: (1) \textbf{No RAG}: the model relies solely on parametric knowledge, (2) \textbf{BM25}: lexical retrieval over English Wikipedia, and (3) \textbf{Embedding}: semantic retrieval using multilingual embeddings with FAISS. This yields 6 configurations per model size (18 total). The resulting factorial design isolates the contributions of model scale, reasoning, and retrieval to overall and rare-entity accuracy.

\paragraph{Baselines.} We compare against four baselines in total. Three published baselines on MERLIN:
\textbf{GEMEL}~\cite{shi-etal-2024-generative} (58.7\%), a generative multimodal entity linking approach;
\textbf{mGENRE}~\cite{decao2021multilingualautoregressiveentitylinking} (72.9\%), which performs multilingual autoregressive entity retrieval with constrained beam search over Wikipedia titles; and
\textbf{Cultural Pangea}~\cite{nyandwi2025groundingmultilingualmultimodalllms} (81.1\%), the current SOTA on MERLIN.

We also create an additional retrieval-aware baseline using our embedding retrieval. Since CulturalPangea does not support tool calling, \textbf{CulturalPangea-RAG} prepends the top-5 retrieved (title, description) pairs to Pangea's input.
\section{Results}
\label{sec:results}

We evaluate our framework on both the full MERLIN test set and the rare entity slices defined in \S\ref{sec:rarity}. We then address three research questions: (RQ1) How does the advantage over existing methods scale with entity rarity? (RQ2) What drives the gains on rare entities: reasoning, retrieval, or their combination? (RQ3) Can smaller reasoning models with retrieval match larger ones?

% ===================================================
% MAIN RESULTS
% ===================================================

\subsection{Main Results}
\label{subsec:main-results}

Table~\ref{tab:main-results} presents accuracy on the full MERLIN test set. Our best system, 8B-Think+Embed, achieves 87.9\% average accuracy, outperforming Cultural Pangea by \textbf{+6.9\%}, with gains of +10.0\% on Hindi and Indonesian. Reasoning models consistently beat their instruct counterparts under retrieval; 4B-Think+Embed (83.7\%) matches 8B-Instr (83.5\%) with half the parameters, though it uses $\sim$2.8$\times$ more inference tokens than 8B-Instr no-RAG (Appendix~\ref{app:cost}, Table~\ref{tab:cost-breakdown}). The trade is fewer parameters for more compute per example.

The advantage remains 4.9\% under redirect-aware scoring (Appendix~\ref{app:redirect-eval}).

% ===================================================
% RARE ENTITY RESULTS — HERO FINDING
% ===================================================

\paragraph{RQ1: How Does the Advantage Scale with Entity Rarity?}
\label{subsec:rare-results}

Across all 15 rare-entity slices, gains range from +5.5\% to +23.3\%. The largest gains occur for qualifiers (+23.3\%), statements (+22.1\%), and Wikidata outgoing links (+21.7\%), compared with +6.9\% on the full dataset. Fourteen of the 15 rare-entity gains exceed the full-dataset gain. Appendix~\ref{app:full-rare-slices} reports all 15 slices.

The largest gains occur on Wikidata structural metrics, especially qualifiers, statements, and outgoing links (Table~\ref{tab:rare-advantage}).

\begin{table}[h]
\centering
\small
\begin{tabular}{lcc}
\toprule
\textbf{Split (Bottom 5\%)} & \textbf{Ours} & \textbf{$\Delta$ vs Pangea} \\
\midrule
Full dataset & 87.9 & +6.9 \\
\midrule
\multicolumn{3}{l}{\textit{Wikidata structural}} \\
\quad WD Out-Links & 66.3 & +21.7 \\
\quad Statements & 66.1 & +22.1 \\
\quad Qualifiers & 68.4 & \textbf{+23.3} \\
\quad Lang. Editions & 63.9 & +16.3 \\
\quad WD In-Links & 69.0 & +11.5 \\
\midrule
\multicolumn{3}{l}{\textit{Wikipedia engagement}} \\
\quad Pageviews & 57.5 & +14.1 \\
\quad Categories & 59.5 & +18.3 \\
\bottomrule
\end{tabular}
\caption{8B-Think+Embed accuracy (\%) and advantage over Pangea on rare entity slices. The advantage reaches +23.3\% on rare entity slices, compared with +6.9\% on the full dataset.}
\label{tab:rare-advantage}
\end{table}

% ===================================================
% RQ2 — RETRIEVAL CONTRIBUTION
% ===================================================

\paragraph{RQ2: What Drives the Gains: Reasoning, Retrieval, or Their Combination?}
\label{subsec:rag-rarity}

Table~\ref{tab:rag-amplification} reveals that the benefit of retrieval grows dramatically on rare entities. For 8B-Think+Embed, the RAG delta grows from +3.8\% on the full dataset to +18.8\% on language editions, a 5.0$\times$ increase. On common entities, the model may already know the answer parametrically, so retrieval is redundant. On rare entities, parametric knowledge fails and retrieval becomes essential.

\begin{table}[h]
\centering
\small
\begin{tabular}{lccc}
\toprule
\textbf{Split} & \textbf{Th+Em} & \textbf{Th+BM} & \textbf{In+BM} \\
\midrule
Full dataset & +3.8 & +1.7 & $-4.9$ \\
\midrule
Lang.\ Ed. & +18.8 & +12.2 & +12.6 \\
Statements & +16.6 & +12.1 & +10.1 \\
WD Out-Links & +15.8 & +12.1 & +9.0 \\
Qualifiers & +15.7 & +12.5 & +8.1 \\
Categories & +8.6 & +2.0 & +0.4 \\
\bottomrule
\end{tabular}
\caption{RAG delta (\%) vs.\ no-RAG baseline computed as per-language macro-average. Th+Em = 8B-Think+Embed, Th+BM = 8B-Think+BM25, In+BM = 8B-Inst+BM25.}
\label{tab:rag-amplification}
\end{table}

\paragraph{The Instruct Reversal.}
On the full dataset, BM25 retrieval hurts the instruct model by $-4.9\%$ (Table~\ref{tab:rag-amplification}). Yet on the structural rare-entity slices shown in Table~\ref{tab:rag-amplification}, the same BM25 retrieval helps by +8.1\% to +12.6\%. This reversal occurs because instruct models issue searches 3.2-3.7 searches per example with no deliberation between them, flooding their context with retrieved results they cannot effectively filter. On common entities this noise overwhelms the correct parametric answer. On rare entities any retrieval signal helps because parametric knowledge is absent. Their queries degrade over successive iterations, with verbatim repetition climbing to 34\% by search number 15+ (Appendix~\ref{app:search-behavior}). Thinking models avoid all of this. Think+BM25 shows consistent positive deltas on both full (+1.7\%) and rare (+12\%) data.

\paragraph{Reasoning alone does not help on rare entities.}
The reasoning effect (Think vs.\ Inst, both without RAG) is not significant on any rare split ($p > 0.5$ across all slices; Table~\ref{tab:stat-tests}). Think and Inst models perform similarly on rare splits without retrieval. The main advantage comes from the combination of reasoning and retrieval. Reasoning enables effective use of retrieved information, while retrieval provides information that reasoning alone lacks.

\paragraph{How reasoning models use retrieval differently.}
Thinking models make fewer but more deliberate searches (1.0--2.2 as opposed to 3.2-5.2 per example for instruct models). They also generate 1,454--1,534 tokens between consecutive searches. This deliberation manifests in query strategy.
56--58\% of 8B-Think model's search transitions are refinements that add disambiguation context to the
previous query, compared to 35--40\% for Instruct variant (Appendix~\ref{app:query-content}). 

% ===================================================
% SCALING ANALYSIS
% ===================================================

\paragraph{RQ3: Can Smaller Reasoning Models Match Larger Ones?}
\label{subsec:scaling}
Accuracy across model sizes (Figure~\ref{fig:scaling}, Appendix~\ref{app:reasoning-vs-size}) shows two patterns. First, Embed-Think leads at 4B and 8B. At 4B it already surpasses 8B-Instr, however at 2B the no-RAG baselines outperform all retrieval-augmented configurations. Further analysis reveals that 2B-Think issues only one (forced) search on 97\% of examples. It cannot effectively use the tool at all, indicating that effective tool use is an emergent capability requiring sufficient model scale (Appendix~\ref{app:search-behavior}). Second, the Think--Instruct gap is substantially wider under retrieval (both BM25 and embedding) than without, suggesting that reasoning and retrieval are complementary.

4B-Think+Embed and 8B-Inst are virtually identical on the full dataset (+0.3\%), but on rare entities the smaller reasoning model wins by +5 to +7\% (Appendix~\ref{app:reasoning-vs-size}). This demonstrates that reasoning with retrieval compensates for pure model size.

% ===================================================
% ERROR ANALYSIS
% ===================================================

\subsection{Error Analysis}
\label{subsec:errors}

We analyzed 8B-Think+Embed's 841 errors (Table~\ref{tab:error-taxonomy}) to provide a taxonomy and inform future work. Further error analysis can be found in Appendix~\ref{app:error-shift} and \ref{app:pipeline}. Detailed search behavior analysis, computational cost breakdowns, and query content analysis are in Appendices~\ref{app:search-behavior}--\ref{app:query-content}.

\begin{table}[t]
\centering
\small
\begin{tabular}{lrr}
\toprule
\textbf{Error Category} & \textbf{N} & \textbf{\%} \\
\midrule
Completely Wrong & 436 & 51.8 \\
Name Format & 180 & 21.4 \\
Disambiguation (Specificity) & 112 & 13.3 \\
Wikipedia Variant & 55 & 6.5 \\
Concept Granularity & 45 & 5.4 \\
Empty (Pipeline Error) & 13 & 1.5 \\
\midrule
Total & 841 & 100.0 \\
\bottomrule
\end{tabular}
\caption{Error taxonomy for 8B-Think+Embed on the full MERLIN test set across all languages.}
\label{tab:error-taxonomy}
\end{table}

We additionally decompose
errors by pipeline stage
(Appendix~\ref{app:pipeline}). The dominant bottleneck is
retrieval: in 72\% of errors, the correct entity was never
surfaced by search. Even the deliberate query refinement
that Thinking models employ
(Appendix~\ref{app:query-content}) cannot always bridge
cross-lingual gaps between non-Latin mentions and English
Wikipedia titles. A further 23.5\% of errors involve the
model engaging with the correct entity in its reasoning but
ultimately rejecting it, indicating that disambiguation
remains an open challenge even for reasoning-native models. Across configurations
(Appendix~\ref{app:error-shift}), Think+Embed has the
fewest total errors (841 vs.\ 1,493 for 8B-Inst+BM25) but
the highest proportion of Completely Wrong cases. Retrieval
and reasoning resolve the easier categories, thus concentrating
residual errors on genuinely hard cases.

\paragraph{Worked examples.}
\begin{compactenum}
\item Japanese mention \jp{米} (Bei): a newspaper abbreviation of \jp{米国} (Beikoku, ``the United States''). The model read it as a surname, searched ``Mi surname,'' and predicted \textit{Mi (surname)}; ``United States'' was never queried. The same pattern recurs for \jp{英} (``the United Kingdom''). Category: completely wrong (retrieval failure).
\item Hindi mention \dev{i-lAEmk pr\2prA} (``Islamic tradition''): search returned the gold title \textit{Islamic culture} at rank 2, but the model answered \textit{Islamic family law}, a neighboring concept in the same domain, not the target. Category: concept granularity.
\item Indonesian mention Kinabalu: the search ``Kinabalu'' retrieved the gold \textit{Kinabalu (federal constituency)} at rank 9, while the city \textit{Kota Kinabalu} sat at rank 1; the model chose the city. Category: disambiguation.
\end{compactenum}

\paragraph{Errors on head versus rare entities.}
An entity is rare if it falls in the bottom 5\% of any of the 15 rarity metrics (\S\ref{sec:rarity}), and head otherwise. The error rate is far higher on rare than head entities: 31.1\%, versus 8.4\%. The error mix also shifts significantly ($\chi^2 = 40.84$, $p = 1.0\times10^{-7}$). As shown in Table~\ref{tab:head-vs-rare}, disambiguation errors become less common on rare entities, while concept-granularity and name-format errors become more common.

\begin{table}[h]
\centering
\footnotesize
\setlength{\tabcolsep}{4pt}
\begin{tabular}{lcc}
\toprule
\textbf{Error Category} & \textbf{Head} & \textbf{Rare} \\
\midrule
Completely Wrong & 48.9\% & 56.1\% \\
Name Format & 20.3\% & 23.0\% \\
Disambiguation & 16.3\% & 9.0\% \\
Wikipedia Variant & 9.1\% & 2.9\% \\
Concept Granularity & 3.0\% & 8.7\% \\
Empty & 2.4\% & 0.3\% \\
\bottomrule
\end{tabular}
\caption{Share of errors (\%) by category, for 8B-Think+Embed, on head entities versus rare entities. Rare-entity errors shift toward completely-wrong and concept-granularity and name-format, away from disambiguation.}
\label{tab:head-vs-rare}
\end{table}

\paragraph{Effect of Writing Script.}
\label{subsec:script}

We compare Latin-script mentions (Indonesian, Vietnamese) with non-Latin ones (Hindi, Tamil, Japanese) for our best embedding system (Table~\ref{tab:script-strat}). First-search hit rates are similar across scripts, but retrieval failure is a larger share of errors on non-Latin inputs, and embedding's advantage over BM25 is largest on non-Latin inputs and on rare subsets.

\begin{table}[h]
\centering
\footnotesize
\setlength{\tabcolsep}{4pt}
\begin{tabular}{lcc}
\toprule
\textbf{Measure (Embed)} & \textbf{Latin} & \textbf{Non-Latin} \\
\midrule
First-search hit rate & 37.0\% & 39.2\% \\
Retrieval failure share of errors & 62.5\% & 76.8\% \\
Gain over BM25 (full set) & +1.9\% & +2.2\% \\
Gain over BM25 (rare subsets) & +6.3\% & +6.9\% \\
\bottomrule
\end{tabular}
\caption{Embedding retrieval statistics for Latin-script (Id, Vi) versus non-Latin-script (Hi, Ta, Ja) inputs, 8B-Think+Embed.}
\label{tab:script-strat}
\end{table}

% \paragraph{Error taxonomy.} Table~\ref{tab:error-taxonomy}
% categorizes 8B-Think+Embed's 841 errors. Notably, retrieval
% resolves the easier categories (Name Format, Wikipedia
% Variant), so Think+Embed has the fewest total errors but the
% highest proportion of Completely Wrong
% (Appendix~\ref{app:error-shift}).

% \paragraph{Pipeline decomposition.} Of 841 errors, 72.1\%
% are retrieval failures (correct entity never retrieved),
% 23.5\% extraction failures (correct entity retrieved but
% wrong title generated), and 2.9\% selection failures.
% Thinking models mitigate retrieval failure through
% deliberate query refinement---58\% of search transitions
% add disambiguation context, versus 21--40\% for Instruct
% models, whose queries degrade toward verbatim repetition
% (Appendix~\ref{app:query-content})---but hit a
% cross-lingual ceiling when non-Latin mentions diverge from
% English Wikipedia titles. Extraction failures suggest
% constrained decoding over retrieved titles as low-hanging
% fruit.

\begin{table}[h]
\centering
\small
\begin{tabular}{lcr}
\toprule
\textbf{Split} & \textbf{$\Delta$ (\%)} & \textbf{$p$-value} \\
\midrule
\multicolumn{3}{l}{\textit{RAG effect (Think+Embed vs Think)}} \\
\quad Lang.\ Ed. & +18.8 & $<$0.001 \\
\quad Statements & +16.6 & $<$0.001 \\
\quad WD Out-Links & +15.8 & $<$0.001 \\
\midrule
\multicolumn{3}{l}{\textit{Reasoning effect (Think vs Inst, no RAG)}} \\
\quad Lang.\ Ed. & $-1.4$ & 0.575 \\
\quad Statements & +0.9 & 0.766 \\
\quad WD Out-Links & +0.3 & $>0.999$ \\
\bottomrule
\end{tabular}
\caption{Statistical significance (McNemar's and bootstrap tests) computed on examples pooled across languages. RAG effects are significant on all rare splits. Reasoning alone has no significant effect.}
% \vspace{-0.5em}
\label{tab:stat-tests}
\end{table}

\section{Conclusion}
We showed that unpopularity is not the same as rarity. Entities with sparse Wikidata structure and few Wikipedia language editions cause a more severe performance drop than pageview-based analyses would suggest. This indicates that prior work has underestimated the difficulty of the cultural long tail.

Our simple framework, in which a reasoning-capable VLM iteratively searches and reasons over Wikipedia, achieves +6.9\% over the state of the art on MERLIN and up to +23.3\% on the hardest rare-entity slices. The largest gains come from combining reasoning and retrieval. Reasoning without retrieval has no significant effect on rare entities. Retrieval can help rare entities without reasoning, but pairing it with reasoning produces the strongest overall system. For the 8B reasoning model, the retrieval gain increases from +3.8\% on the full dataset to +18.8\% on structurally sparse entities, a 5.0$\times$ increase. A 4B reasoning model with retrieval matches an 8B instruct model overall and outperforms it by +5 to +7\% on rare entities, suggesting an alternative to scaling.

Retrieval failure accounts for 72\% of residual errors, making cross-lingual retrieval a useful direction for future work. Even with retrieval, disambiguation on the long tail still remains an open problem for reasoning models as well.

\newpage

\section*{Limitations}
Our controlled factorial experiments use Qwen3-VL~\cite{qwen3technicalreport}, chosen because it supports comparisons across reasoning variants and scales (Section~\ref{sec:method}). The GLM check in Appendix~\ref{app:second-family} shows that adding retrieval in thinking mode improves all 15 rare-entity slices. However, GLM's non-thinking mode could not sustain the retrieval loop. Thus, the cross-family evidence supports the rare-entity retrieval benefit but does not establish that the full reasoning-by-retrieval interaction generalizes across model families.

We evaluate on five languages from the MERLIN benchmark and target English Wikipedia as the knowledge base. All MERLIN entities have English articles by construction, but this constraint limits applicability to entities without English coverage. Generalization to languages with even sparser Wikipedia representation, such as African languages, remains unknown and would test the limits of retrieval-augmented approaches for the cultural long tail.

Retrieval is the dominant bottleneck in our pipeline, accounting for 72\% of errors. Embedding retrieval partially addresses cross-lingual mismatch for non-Latin scripts, but neither retrieval method guarantees recall when transliterations diverge substantially between the source language and English Wikipedia titles. Improving cross-lingual retrieval in the context of LLM tool use remains the most impactful direction for future work.

We follow MERLIN's standard exact-match evaluation
protocol, which penalizes predictions that produce valid but non-canonical title strings (e.g., redirects or common abbreviations) identically to genuinely incorrect predictions. This is consistent with all prior work on MERLIN and ensures comparability across systems.

\section*{Ethical Considerations}

This work uses publicly available data: the MERLIN benchmark~\cite{ramamoorthy2025merlintestbedmultilingualmultimodal}, Wikipedia, and Wikidata. No personal data or human subjects are involved. Our system inherits biases present in Wikipedia's coverage, which is known to underrepresent non-Western entities and perspectives~\cite{Hecht_2010}. While our rarity characterization helps surface these gaps, the system itself cannot correct underlying knowledge base biases. We note that entity linking systems, including ours, may perform less reliably on entities from marginalized communities that are systematically underrepresented in Wikipedia.

\section*{Acknowledgments}
We thank Ibrahim AlRayes for his help with this project, and Jean de Dieu
Nyandwi and Zaid Sheikh for sharing resources that supported this work. We
also thank the members of NeuLab for their helpful feedback.

This work was supported in part by a research grant from the
Defence Science and Technology Agency (DSTA), Singapore.
\newpage

\bibliography{custom}

% \newpage

\appendix

\label{sec:app}

\clearpage

\section{Appendix}
\subsection{Rarity Metric Independence}
\label{app:metric-overlap}

\begin{figure*}[t!]
\centering
\includegraphics[width=\textwidth]{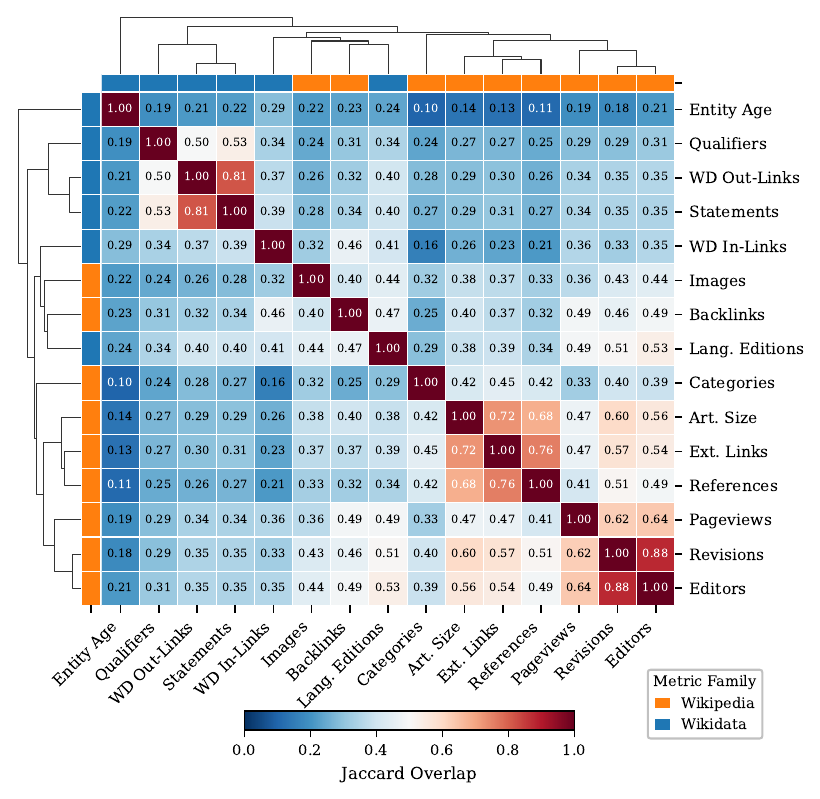}
\caption{Pairwise Jaccard overlap between bottom-5\% rare entity sets for each of the 15 rarity metrics. Mean overlap is 37\%, confirming that different metrics identify substantially different entities as rare. Sidebar color indicates metric family (orange = Wikipedia, blue = Wikidata).}
\label{fig:metric-overlap}
\end{figure*}

To verify that the 15 rarity metrics identify genuinely different entities as ``rare'' rather than measuring a single underlying construct, we compute pairwise Jaccard overlap between the bottom-5\% entity sets for each metric (Figure~\ref{fig:metric-overlap}). The mean pairwise overlap is only 37\%, confirming that different metrics flag largely non-overlapping entity sets. Within-family overlap is moderate (Wikipedia: 48\%, Wikidata: 38\%), while between-family overlap is lower still (30\%), with some cross-family pairs sharing as few as 10\% of entities.

Qualitative analysis reinforces this finding. We identify 50 entities that are Wikidata-rare (bottom-5\% on $\geq$3 Wikidata metrics) but Wikipedia-normal (zero Wikipedia metrics flagged), including well-documented news events like the \emph{2016 Indian banknote demonetisation} (36K pageviews, 980 editors, but only 11 Wikidata statements) and the \emph{Pittsburgh synagogue shooting} (88K pageviews but structurally sparse in Wikidata). Conversely, 92 entities are Wikipedia-rare but Wikidata-normal, such as the \emph{Muttahida Qaumi Movement} (minimal English Wikipedia article but 287 Wikidata incoming links). These examples illustrate that editorial attention and structural connectivity are different dimensions of rarity.

\subsection{Baseline Degradation Across All Systems}
\label{app:baseline-rarity}

\begin{table}[h]
\centering
\small
\begin{tabular}{lrrr}
\toprule
\textbf{Eval.\ Slice} & \textbf{GEMEL} & \textbf{mGENRE} & \textbf{Pangea} \\
\midrule
Full Test Set & 58.7 & 72.9 & 81.1 \\
\midrule
\multicolumn{4}{l}{\textit{Wikidata structural (Bottom 5\%)}} \\
\quad Lang. Editions & 32.3 & 40.7 & 47.6 \\
\quad Statements & 31.9 & 42.7 & 44.0 \\
\quad WD Out-Links & 31.3 & 42.4 & 44.6 \\
\quad Qualifiers & 32.4 & 44.9 & 45.1 \\
\quad WD In-Links & 37.2 & 45.2 & 57.5 \\
\midrule
\multicolumn{4}{l}{\textit{Wikipedia engagement (Bottom 5\%)}} \\
\quad Pageviews & 29.0 & 42.6 & 43.3 \\
\quad Categories & 25.4 & 44.6 & 41.2 \\
\bottomrule
\end{tabular}
\caption{Baseline accuracy (\%) on the full test set and bottom-5\% rare entity slices. All three baselines degrade on rare entities.}
% \vspace{-1em}
\label{tab:baseline-all-rarity}
\end{table}

% ===================================================
% APPENDIX B: RAG DELTA BY SIZE
% ===================================================

\begin{figure*}[t]
\centering
\includegraphics[width=\textwidth]{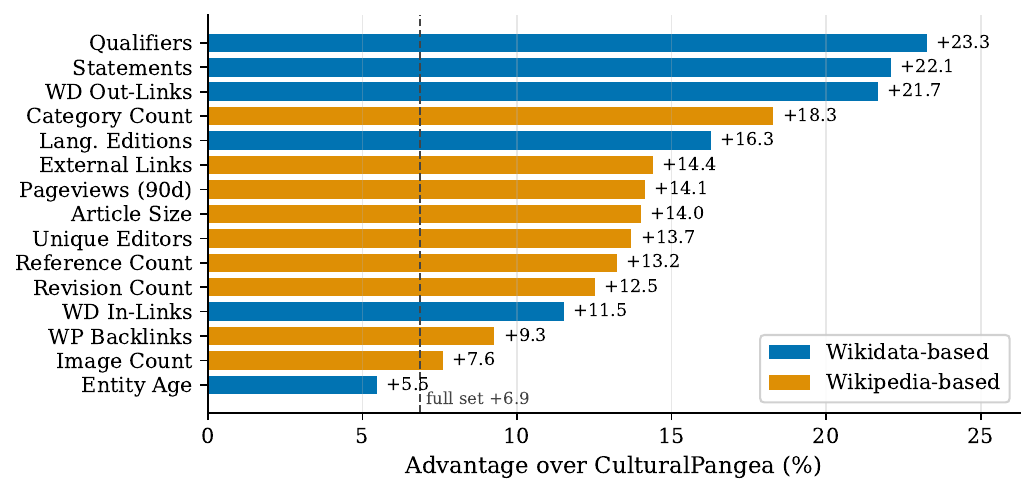}
\caption{Advantage (\%) of 8B-Think+Embed over Cultural Pangea across all 15 bottom-5\% rare entity slices, sorted by magnitude.}
\label{fig:advantage-growth-appendix}
\end{figure*}

\subsection{Full Rare Entity Slice Breakdown}
\label{app:full-rare-slices}

Figure~\ref{fig:advantage-growth-appendix} presents the advantage of 8B-Think+Embed over Cultural Pangea across all 15 bottom-5\% rare entity slices. Every gain is positive, and the largest gains occur for qualifiers, statements, and Wikidata outgoing links.

\subsection{Redirect-Aware Evaluation}
\label{app:redirect-eval}

Our main results use MERLIN's standard exact-match protocol. We additionally evaluate after resolving English Wikipedia redirects. Redirect-aware scoring raises our accuracy from 87.9\% to 89.5\% and CulturalPangea's accuracy from 80.1\% to 84.6\%. Our advantage remains 4.9\% (paired bootstrap 95\% CI [+4.01, +5.79]), showing that the improvement is not caused by non-canonical title variants. We keep exact match as the primary metric for comparability with prior MERLIN work.

\begin{table}[h]
\centering
\small
\begin{tabular}{lrrr}
\toprule
\textbf{Scoring} & \textbf{Ours} & \textbf{Pangea} & \textbf{Difference} \\
\midrule
Exact match & 87.9 & 80.1 & +7.8 \\
Redirect-aware & 89.5 & 84.6 & +4.9 \\
\bottomrule
\end{tabular}
\caption{Accuracy (\%) under exact-match and redirect-aware scoring.}
\label{tab:redirect-eval}
\end{table}

\subsection{RAG Effect by Model Size}
\label{app:rag-by-size}

Table~\ref{tab:rag-by-size} shows the Think-vs-Instruct gap broken down by model size and retrieval method. The gap is largest with embedding retrieval, consistent across scales.

\begin{table}[h]
\centering
\small
\begin{tabular}{lccc}
\toprule
\textbf{Size} & \textbf{No RAG} & \textbf{BM25} & \textbf{Embed} \\
\midrule
2B & $-0.8$ & +6.8 & +4.8 \\
4B & +3.4 & +8.0 & +7.6 \\
8B & +0.7 & +7.3 & +5.0 \\
\bottomrule
\end{tabular}
\caption{Think $-$ Instruct gap (\%, average across languages) by model size and retrieval method. Reasoning models benefit more from retrieval at every scale.}
\label{tab:rag-by-size}
\end{table}

% ===================================================
% APPENDIX C: ROBUSTNESS ANALYSIS
% ===================================================

\subsection{Robustness Analysis}
\label{app:robustness}

We define the \emph{robustness gap} as the difference between Pangea's accuracy drop and our system's accuracy drop on each rare slice. A positive robustness gap means our system degrades less.

Table~\ref{tab:robustness} presents the full robustness analysis. Our system degrades less than Pangea on 14 of the 15 rare entity slices, with robustness gaps up to 16.4\%. Entity age is the single exception.

\begin{table}[h]
\centering
\small
\begin{tabular}{lrrr}
\toprule
\textbf{Metric} & \textbf{Ours} & \textbf{Pangea} & \textbf{Gap} \\
\midrule
Qualifiers & $-19.5$ & $-35.9$ & 16.4 \\
Statements & $-21.8$ & $-37.0$ & 15.2 \\
WD Out-Links & $-21.7$ & $-36.5$ & 14.8 \\
Categories & $-28.4$ & $-39.9$ & 11.4 \\
Lang.\ Editions & $-24.0$ & $-33.4$ & 9.4 \\
Ext.\ Links & $-29.6$ & $-37.2$ & 7.5 \\
Pageviews & $-30.5$ & $-37.7$ & 7.3 \\
Art.\ Size & $-26.9$ & $-34.0$ & 7.1 \\
Editors & $-29.7$ & $-36.5$ & 6.9 \\
References & $-27.7$ & $-34.1$ & 6.4 \\
Revisions & $-30.7$ & $-36.4$ & 5.7 \\
WD In-Links & $-18.9$ & $-23.6$ & 4.7 \\
WP Backlinks & $-28.7$ & $-31.1$ & 2.4 \\
Images & $-26.5$ & $-27.3$ & 0.7 \\
Entity Age & $-16.8$ & $-15.4$ & $-1.4$ \\
\bottomrule
\end{tabular}
\caption{Accuracy drop (\%) from full dataset to bottom-5\% slices. ``Gap'' = Pangea drop $-$ our drop. Sorted by robustness gap.}
\label{tab:robustness}
\end{table}

% ===================================================
% APPENDIX: THRESHOLD ROBUSTNESS
% ===================================================

\subsection{Threshold Robustness}
\label{app:threshold-robustness}

Table~\ref{tab:threshold-robustness} shows that our system's advantage over Pangea remains positive across 1\%, 5\%, and 10\% rarity thresholds. The gap ranges from +8.0\% to +41.1\%, confirming that the result is not caused by the 5\% cutoff.

\begin{table}[t]
\centering
\small
\setlength{\tabcolsep}{3pt}
\begin{tabular}{llrrrr}
\toprule
\textbf{Metric} & \textbf{Thr.} & \textbf{$n$} & \textbf{Ours} & \textbf{Pangea} & \textbf{Gap} \\
\midrule
\multirow{3}{*}{Lang.\ Ed.}
  & 1\% & 9   & 55.6 & 22.2 & +33.3 \\
  & 5\% & 346 & 63.9 & 47.6 & +16.3 \\
  & 10\% & 548 & 66.6 & 51.5 & +15.1 \\
\midrule
\multirow{3}{*}{Statements}
  & 1\% & 35  & 66.1 & 30.9 & +35.3 \\
  & 5\% & 339 & 66.1 & 44.0 & +22.1 \\
  & 10\% & 573 & 68.2 & 49.2 & +19.0 \\
\midrule
\multirow{3}{*}{WD Out-Links}
  & 1\% & 24  & 65.6 & 24.4 & +41.1 \\
  & 5\% & 331 & 66.3 & 44.6 & +21.7 \\
  & 10\% & 579 & 68.0 & 49.4 & +18.6 \\
\midrule
\multirow{3}{*}{Qualifiers}
  & 1\% & 15  & 66.7 & 46.7 & +20.0 \\
  & 5\% & 323 & 68.4 & 45.1 & +23.3 \\
  & 10\% & 551 & 72.1 & 51.4 & +20.7 \\
\midrule
\multirow{3}{*}{Pageviews}
  & 1\% & 34  & 44.2 & 36.2 & +8.0 \\
  & 5\% & 350 & 57.5 & 43.3 & +14.1 \\
  & 10\% & 549 & 61.0 & 47.3 & +13.8 \\
\bottomrule
\end{tabular}
\caption{Accuracy (\%) at 1\%, 5\%, and 10\% rarity thresholds for five metrics, using per-language macro-averages. The 5\% rows reproduce Table~\ref{tab:rare-advantage}. Gains are positive for every metric and threshold, ranging from +8.0\% to +41.1\%.}
\label{tab:threshold-robustness}
% \vspace{-1em}
\end{table}

\subsection{Accuracy by Rarity Decile}
\label{app:deciles}

\begin{figure}[t]
\centering
\includegraphics[width=0.47\textwidth]{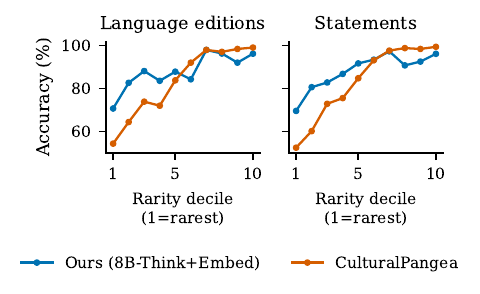}
\caption{Accuracy (\%) by rarity decile (1 = rarest, 10 = most common) for two representative Wikidata-structural metrics. Each decile is a disjoint bin of one tenth of the entities. Our system (8B-Think+Embed) degrades roughly half as steeply as CulturalPangea toward the sparse end.}
\label{fig:decile-curves}
\end{figure}

Figure~\ref{fig:decile-curves} additionally plots accuracy against rarity decile (1 = rarest, 10 = most common) for two representative Wikidata-structural metrics. Each decile is a disjoint bin holding one tenth of the test entities, ranked by the metric, so the point at decile $d$ is the accuracy computed on that bin alone rather than a cumulative accuracy over all entities up to $d$. This makes the plot a conditional accuracy curve, showing how accuracy varies with rarity level instead of tracking a running total. Read this way, the baseline declines steeply and near-monotonically toward the sparse end on every Wikidata-structural metric (Spearman $\rho \geq 0.92$), while our system degrades roughly half as fast (decile-1-vs-10 gap $\approx$ 26\% vs.\
45--47\%). At the common-entity end (decile 10) the baseline is at ceiling and edges ahead of us by 2 to 3\%, but the two curves cross as entities get rarer and the gap opens in our favor. The advantage is therefore a gradient across the full distribution, not an artifact of any single threshold, and the small head-of-distribution deficit is outweighed by our gains on the other nine deciles (net +6.9\% overall).

% ===================================================
% APPENDIX E: ERROR TAXONOMY SHIFT ON RARE ENTITIES
% ===================================================

\begin{figure*}[t]
\centering
\includegraphics[width=0.8\textwidth]{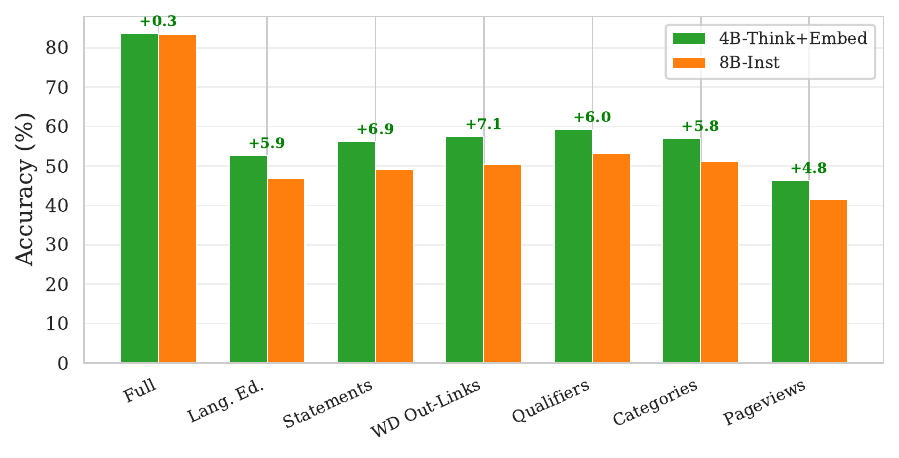}
\caption{4B-Think+Embed vs.\ 8B-Inst on the full dataset and rare entity slices. The two systems are nearly identical on the full dataset but diverge on rare entities, with the smaller reasoning model winning by +5--7\%.}
\label{fig:reasoning-vs-size}
\end{figure*}

\subsection{Error Taxonomy Shift on Rare Entities}
\label{app:error-shift}

% \begin{figure*}[h]
% \centering
% \includegraphics[width=\textwidth]{sections/figures/rare_fig_r4_error_type_shift.pdf}
% \caption{Error type distribution on full dataset vs.\ rare entity slices. On rare entities, ``Completely Wrong'' increases while ``Disambiguation'' decreases, reflecting a shift from disambiguating among known entities to failing to find the entity at all.}
% \label{fig:error-shift}
% \end{figure*}

Table~\ref{tab:error-full} presents the full error taxonomy across all configurations.

\begin{table*}[h]
\centering
\small
\begin{tabular}{lrrrrr}
\toprule
\textbf{Error Category} & \textbf{8B-Th+Em} & \textbf{8B-Th+BM} & \textbf{8B-Think} & \textbf{8B-Inst} & \textbf{8B-In+BM} \\
\midrule
Completely Wrong & 436 (51.8\%) & 501 (50.8\%) & 448 (40.5\%) & 508 (44.0\%) & 560 (37.5\%) \\
Name Format & 180 (21.4\%) & 221 (22.4\%) & 333 (30.1\%) & 326 (28.2\%) & 362 (24.2\%) \\
Disambiguation & 112 (13.3\%) & 134 (13.6\%) & 121 (10.9\%) & 119 (10.3\%) & 349 (23.4\%) \\
Wikipedia Variant & 55 (6.5\%) & 73 (7.4\%) & 130 (11.8\%) & 121 (10.5\%) & 61 (4.1\%) \\
Concept Granularity & 45 (5.4\%) & 44 (4.5\%) & 61 (5.5\%) & 68 (5.9\%) & 48 (3.2\%) \\
Empty (Pipeline Error) & 13 (1.5\%) & 13 (1.3\%) & 13 (1.2\%) & 13 (1.1\%) & 113 (7.6\%) \\
\midrule
Total & 841 & 986 & 1106 & 1155 & 1493 \\
\bottomrule
\end{tabular}
\caption{Full error taxonomy across 8B configurations. Th+Em has the fewest total errors (841) but the highest \emph{proportion} of ``Completely Wrong'' (51.8\%), because its easier errors (Name Format, Wikipedia Variant) are resolved by retrieval, leaving harder cases.}
\label{tab:error-full}
\end{table*}

% ===================================================
% APPENDIX F: PIPELINE DECOMPOSITION
% ===================================================

\begin{table}[h!]
\centering
\small
\begin{tabular}{lrrr}
\toprule
\textbf{Stage} & \textbf{Th+Em} & \textbf{Th+BM} & \textbf{In+BM} \\
\midrule
Retrieval Failure & 72.1\% & 89.5\% & 80.9\% \\
Selection Failure & 2.9\% & 1.4\% & 3.7\% \\
Extraction Failure & 23.5\% & 7.8\% & 7.8\% \\
Empty & 1.5\% & 1.3\% & 7.6\% \\
\bottomrule
\end{tabular}
\caption{Pipeline decomposition of errors for RAG-based 8B configurations.}
\label{tab:pipeline-full}
\end{table}

\subsection{Pipeline Decomposition}
\label{app:pipeline}

Table~\ref{tab:pipeline-full} decomposes errors by pipeline stage for RAG-based configurations. Retrieval failures dominate across all configurations, but 8B-Think+Embed shows the lowest retrieval failure rate (72.1\%) due to embedding retrieval's better cross-lingual recall. BM25-based configurations (8B-Think+BM25, 8B-Inst+BM25) have retrieval failure rates of 89.5\% and 80.9\% respectively.

% ===================================================
% APPENDIX G: 4B vs 8B ON RARE ENTITIES
% ===================================================

\subsection{Reasoning Compensates for Size}
\label{app:reasoning-vs-size}

Figure~\ref{fig:reasoning-vs-size} compares 4B-Think+Embed against 8B-Inst across the full dataset and rare entity slices. On the full MERLIN test set, the two configurations are nearly identical (83.7\% vs.\ 83.5\%, a gap of just +0.3\%), despite the reasoning model having half the parameters. However, the systems diverge sharply on rare entities: 4B-Think+Embed outperforms 8B-Inst by +5.9\% on language editions, +6.9\% on statements, and +7.1\% on Wikidata out-links. This suggests that reasoning with retrieval is a more cost-effective strategy than scaling model size alone, particularly for the long tail where parametric knowledge is insufficient and effective use of retrieved evidence becomes the dominant factor.

\begin{figure}[h]
\centering
% \vspace{-0.5em}
\includegraphics[width=0.47\textwidth]{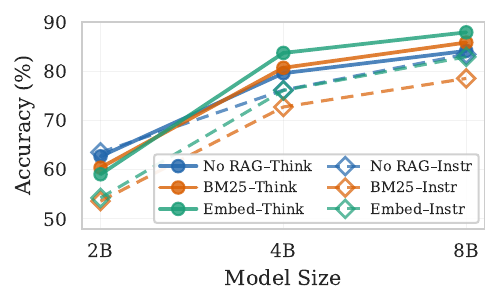}
\caption{Accuracy vs.\ model size (2B/4B/8B) across all 6 configurations.}
\label{fig:scaling}
% \vspace{-0.5em}
\end{figure}

% ===================================================
% APPENDIX H: PROMPTS
% ===================================================

\subsection{Prompt Templates}
\label{app:prompts}

We use the same system prompt across all configurations, varying only the retrieval tool availability. The model receives the article text, image, and marked entity mention, and is instructed to output the English Wikipedia title of the referenced entity. The full prompt is available with our released code on the project page.

\paragraph{System prompt (with retrieval).} The model is instructed that it has access to a Wikipedia search tool. We force the first search call, while all later tool choices are automatic. Each call injects the top-$k$ retrieved titles and descriptions into the context. The model may search up to 20 times per example. The prompt guides a six-step methodology: lock in the entity mention, analyze context for disambiguation, evaluate the image, translate or transliterate for search, search strategically with iterative refinement, and verify the answer. After completing its reasoning, the full trace is passed to a second extraction prompt that instructs the model to output only the final answer. This two-pass design decouples open-ended deliberation from structured answer extraction.

The model also receives an OpenAI-compatible tool definition for \texttt{search\_wikipedia(query, limit)} that instructs it to translate or transliterate entity names to English for best results.

\paragraph{System prompt (without retrieval).}
Identical to the above, but Step 5 is replaced with ``Using your internal knowledge, identify the correct English Wikipedia page title'' and the search tool is not provided.

% ===================================================
% APPENDIX I: SEARCH BEHAVIOR ANALYSIS
% ===================================================

\subsection{Search Behavior Analysis}
\label{app:search-behavior}

We quantify how Thinking and Instruct models differ in their use of the Wikipedia search tool across all 12 RAG configurations (3 sizes $\times$ 2 variants $\times$ 2 retrieval methods).

\paragraph{Search count and deliberation.}
Table~\ref{tab:search-behavior} shows that Instruct models make substantially more searches per example (3.2--5.2 average) than Thinking models (1.0--2.2), yet achieve lower accuracy. Thinking models generate 1,400--3,100 completion tokens between consecutive searches, using this reasoning to analyze retrieved results and decide whether and what to search next. Instruct models generate only 26--30 tokens between searches (bare tool-call overhead) indicating they issue searches with no analysis of previous results. The behavioral contrast is strategic precision vs. brute-force volume. Think models achieve higher accuracy with fewer, more deliberate searches.

\begin{table*}[t]
\centering
\small
\begin{tabular}{llrrrrrr}
\toprule
\textbf{Model} & \textbf{RAG} & \textbf{Avg} & \textbf{Med} & \textbf{\%1} & \textbf{\%5+} & \textbf{\%20} & \textbf{\%Q1} \\
\midrule
2B-Instr & BM25 & 5.2 & 4 & 0 & 31 & 8 & 91 \\
2B-Instr & Embed & 3.3 & 3 & 1 & 8 & 0 & 98 \\
2B-Think & BM25 & 1.0 & 1 & 91 & 0 & 0 & 98 \\
2B-Think & Embed & 1.0 & 1 & 97 & 0 & 0 & 100 \\
\midrule
4B-Instr & BM25 & 3.5 & 3 & 0 & 18 & 1 & 83 \\
4B-Instr & Embed & 3.0 & 3 & 0 & 7 & 0 & 88 \\
4B-Think & BM25 & 1.2 & 1 & 80 & 0 & 0 & 92 \\
4B-Think & Embed & 1.1 & 1 & 93 & 0 & 0 & 97 \\
\midrule
8B-Instr & BM25 & 3.6 & 3 & 0 & 12 & 1 & 95 \\
8B-Instr & Embed & 3.2 & 3 & 0 & 3 & 0 & 98 \\
8B-Think & BM25 & 2.2 & 2 & 29 & 0 & 0 & 74 \\
8B-Think & Embed & 1.5 & 1 & 68 & 0 & 0 & 88 \\
\bottomrule
\end{tabular}
\caption{Search behavior summary across all 12 RAG configurations. ``Avg/Med'': searches per example. ``\%1/\%5+/\%20'': fraction with 1 / 5+ / 20 (max) searches. ``\%Q1'': fraction of searches in first quarter of output.}
\label{tab:search-behavior}
\end{table*}

\begin{figure}[h]
% \vspace{-1em}
\centering
\includegraphics[width=\columnwidth]{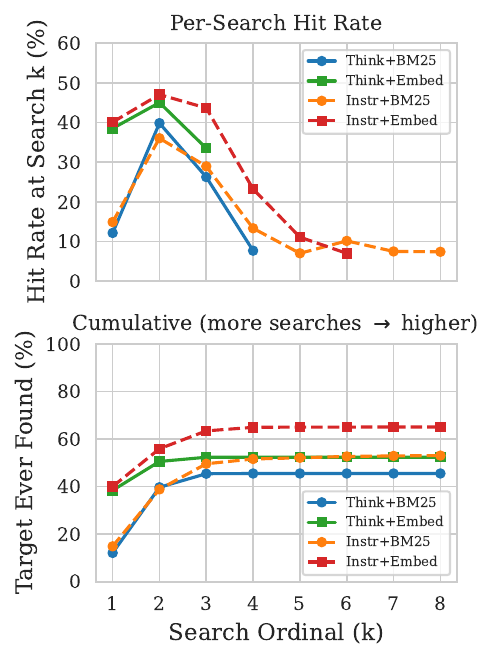}
\caption{Top: per-search hit rate (fraction of examples where search $k$ returns the target). Search 2 is most productive; later searches have sharply diminishing returns. Bottom: cumulative. Instruct (dashed) ends higher only because it makes more total searches, not because any individual search is better. 8B models shown.}
\label{fig:retrieval-success}
% \vspace{-1em}
\end{figure}

% \begin{table*}[t]
% \centering
% \small
% \begin{tabular}{llrrrrrr}
% \toprule
% \textbf{Model} & \textbf{RAG} & \textbf{Jaccard} & \textbf{EditDist} & \textbf{Bigrams} & \textbf{Inter-Tok} & \textbf{Succ@1} & \textbf{Cum@3} \\
% \midrule
% 2B-Instr & BM25 & 0.43 & 0.31 & 0.59 & 29 & 6.7 & 15.0 \\
% 2B-Instr & Embed & 0.50 & 0.36 & 0.69 & 29 & 25.0 & 33.3 \\
% 2B-Think & BM25 & 0.99 & 0.94 & 0.99 & 3092 & 6.2 & 6.8 \\
% 2B-Think & Embed & 0.84 & 0.75 & 1.00 & 3077 & 25.2 & 25.3 \\
% \midrule
% 4B-Instr & BM25 & 0.57 & 0.38 & 0.80 & 30 & 10.8 & 45.9 \\
% 4B-Instr & Embed & 0.58 & 0.39 & 0.81 & 30 & 30.9 & 59.4 \\
% 4B-Think & BM25 & 0.65 & 0.42 & 0.91 & 1974 & 11.5 & 16.7 \\
% 4B-Think & Embed & 0.68 & 0.44 & 0.91 & 1963 & 37.1 & 39.8 \\
% \midrule
% 8B-Instr & BM25 & 0.64 & 0.43 & 0.83 & 27 & 14.9 & 49.6 \\
% 8B-Instr & Embed & 0.65 & 0.45 & 0.84 & 26 & 40.0 & 63.4 \\
% 8B-Think & BM25 & 0.57 & 0.39 & 0.82 & 1534 & 12.1 & 45.4 \\
% 8B-Think & Embed & 0.60 & 0.42 & 0.84 & 1454 & 38.4 & 52.4 \\
% \bottomrule
% \end{tabular}
% \caption{Query diversity and retrieval success. Jaccard/EditDist/Bigrams: mean diversity between consecutive queries (higher=more diverse). Inter-Tok: mean completion tokens between consecutive searches. Succ@1/Cum@3: target found in first search / within first 3 searches.}
% \label{tab:query-diversity}
% \end{table*}

\paragraph{Retrieval success.}
Figure~\ref{fig:retrieval-success} shows per-search and cumulative retrieval success for 8B models. Embedding retrieval finds the target entity in the first search 25--40\% of the time, versus 6--15\% for BM25. The per-search hit rate peaks at search~2 (the primary refinement step) and drops sharply thereafter, showing diminishing returns from additional searches. The cumulative panel shows that Instruct models end up finding the target in more examples overall, but only because they make more searches, not because any individual search is more effective. Despite this higher cumulative retrieval rate, Instruct models achieve lower accuracy than Think models, indicating that finding the target in results is not the bottleneck, reasoning about the results is.

\begin{figure}[!h]
% \vspace{-1em}
\centering
\includegraphics[width=\columnwidth]{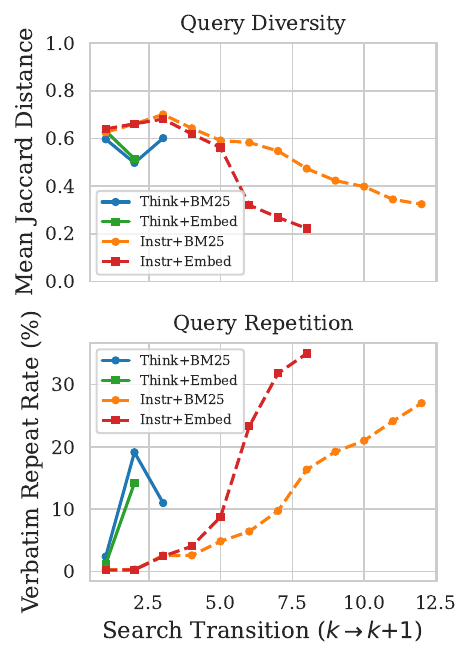}
\caption{Query diversity (left) and verbatim repetition rate (right) across successive search transitions for 8B models. Instruct+BM25 (orange dashed) shows clear degradation: diversity collapses while repetition climbs. Think models (solid) stop after 2--3 transitions with stable quality.}
% \vspace{-1em}
\label{fig:query-evolution}
\end{figure}

\paragraph{Query degradation over successive searches.}
Figure~\ref{fig:query-evolution} reveals that Instruct models' queries degrade over successive iterations. For all models, the first search is a short entity name ($\sim$10 characters) and the second doubles in length by adding disambiguation context. After this, Thinking models stop (68\% of 8B-Think+Embed examples use exactly one search). Instruct models continue with progressively worsening queries: Jaccard diversity drops from 0.63 at transition 1$\to$2 to 0.24 at 19$\to$20, verbatim repetition climbs from 0.3\% to 34\%, and query length bloats from 9.9 to 45.1 characters as the model appends context words. This indicates Instruct models lack an effective stopping criterion.

\begin{figure*}[t]
\centering
\includegraphics[width=\textwidth]{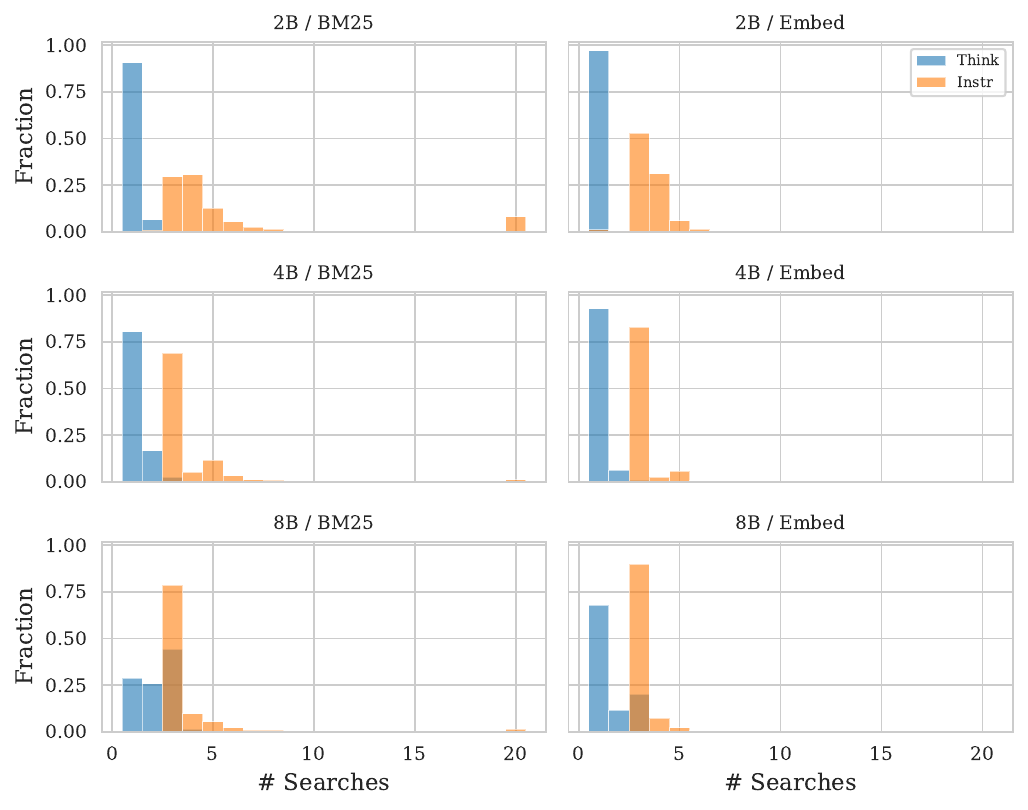}
\caption{Search count distributions across all 12 RAG configurations. Think (blue) concentrates at 1--2 searches; Instruct (orange) spreads across 3--5+. At 2B, Think barely uses search at all.}
\label{fig:search-count-dist}
\end{figure*}

% \begin{figure*}[t]
% \centering
% \includegraphics[width=\textwidth]{sections/figures/retrieval_fig_n2_search_position.pdf}
% \caption{Mean search position distributions across all 12 configurations. Instruct models (orange) cluster near 0. Think models (blue) show broader distributions at 8B, reflecting more spread-out search usage.}
% \label{fig:position-density}
% \end{figure*}

% ===================================================
% APPENDIX J: COMPUTATIONAL COST ANALYSIS
% ===================================================

\subsection{Computational Cost Analysis}
\label{app:cost}

Table~\ref{tab:cost-breakdown} reports the computational cost per example across all 18 configurations, including the 6 no-RAG baselines. All experiments were run on a single NVIDIA L40S GPU (48\,GB VRAM) per configuration, with 10 CPU cores and 25\,GB system RAM. Models were served with SGLang~\cite{zheng2024sglangefficientexecutionstructured}.

\begin{table*}[t]
\centering
\small
\setlength{\tabcolsep}{4pt}
\begin{tabular}{llrrrrrrr}
\toprule
\textbf{Model} & \textbf{RAG} & \textbf{Acc\%} & \textbf{M1 In} & \textbf{M1 Out} & \textbf{M2} & \textbf{Total} & \textbf{Time(s)} & \textbf{Ratio} \\
\midrule
2B-Instr & --- & 63.5 & 1485 & 2856 & 3056 & 7397 & 81.9 & 2.6$\times$ \\
2B-Instr & BM25 & 53.6 & 14961 & 4867 & 3699 & 23527 & 167.8 & 8.4$\times$ \\
2B-Instr & Embed & 54.3 & 10622 & 4048 & 3210 & 17880 & 99.1 & 6.4$\times$ \\
2B-Think & --- & 62.7 & 1487 & 2984 & 1286 & 5757 & 53.3 & 2.0$\times$ \\
2B-Think & BM25 & 60.4 & 3804 & 8514 & 1513 & 13832 & 167.9 & 4.9$\times$ \\
2B-Think & Embed & 59.1 & 3799 & 7396 & 1507 & 12702 & 113.4 & 4.5$\times$ \\
\midrule
4B-Instr & --- & 76.2 & 1485 & 565 & 765 & 2815 & 17.6 & 1.0$\times$ \\
4B-Instr & BM25 & 72.7 & 10145 & 1458 & 367 & 11970 & 66.2 & 4.3$\times$ \\
4B-Instr & Embed & 76.1 & 8906 & 1482 & 391 & 10780 & 51.4 & 3.8$\times$ \\
4B-Think & --- & 79.6 & 1487 & 2272 & 1359 & 5119 & 62.2 & 1.8$\times$ \\
4B-Think & BM25 & 80.7 & 4322 & 5150 & 1327 & 10799 & 166.1 & 3.8$\times$ \\
4B-Think & Embed & 83.7 & 4022 & 4495 & 1320 & 9837 & 124.6 & 3.5$\times$ \\
\midrule
8B-Instr & --- & 83.5 & 1485 & 895 & 1095 & 3475 & 50.7 & 1.2$\times$ \\
8B-Instr & BM25 & 78.6 & 11500 & 502 & 600 & 12602 & 24.6 & 4.5$\times$ \\
8B-Instr & Embed & 82.9 & 9851 & 419 & 536 & 10806 & 19.7 & 3.8$\times$ \\
8B-Think & --- & 84.2 & 1487 & 2027 & 1505 & 5019 & 81.6 & 1.8$\times$ \\
8B-Think & BM25 & 85.8 & 6658 & 5566 & 1486 & 13709 & 247.1 & 4.9$\times$ \\
8B-Think & Embed & 87.9 & 5137 & 4426 & 1489 & 11052 & 176.9 & 3.9$\times$ \\
\bottomrule
\end{tabular}
\caption{Computational cost per example across all 18 configurations. M1 In/Out: Module 1 input/output tokens. M2: Module 2 total tokens. Ratio: number of tokens (cost) relative to cheapest configuration (the most token-efficient).}
\label{tab:cost-breakdown}
\end{table*}

\begin{figure*}[h]
\centering
\includegraphics[width=\textwidth]{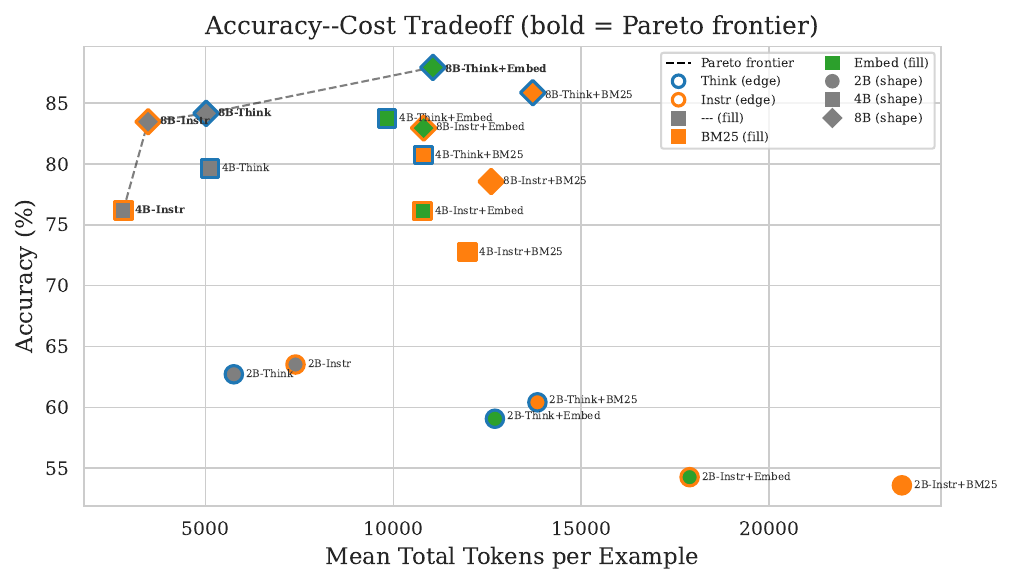}
\caption{Accuracy vs.\ average token cost for all 18 configurations. Bold labels mark the Pareto frontier. Shape encodes model size, edge color encodes Think/Instruct, fill color encodes RAG method.}
\label{fig:pareto}
\end{figure*}

\paragraph{Accuracy--cost tradeoff.}
Figure~\ref{fig:pareto} shows the Pareto frontier of accuracy vs.\ token cost. The frontier runs from 4B-Instr (76.2\%, 2.8k tokens) through 8B-Think (84.2\%, 5.0k tokens) to 8B-Think+Embed (87.9\%, 11.1k tokens). Our best configuration costs 3.9$\times$ the cheapest (4B-Instr), gaining +11.7\% accuracy. The no-RAG 8B-Think baseline (84.2\%, 5.0k tokens) represents a strong low-cost option: 96\% of the best system's accuracy at 45\% of its token cost.

\begin{table*}[t]
\centering
\small
\begin{tabular}{llrrrrr}
\toprule
\textbf{Model} & \textbf{RAG} & \textbf{Tok (F)} & \textbf{Tok (R)} & \textbf{R/F} & \textbf{Dur (F)} & \textbf{Dur (R)} \\
\midrule
2B-Instr & --- & 7397 & 11013 & 1.49 & 81.9 & 133.9 \\
2B-Instr & BM25 & 23527 & 27346 & 1.16 & 167.8 & 178.8 \\
2B-Instr & Embed & 17880 & 19523 & 1.09 & 99.1 & 115.8 \\
2B-Think & --- & 5757 & 6968 & 1.21 & 53.3 & 71.6 \\
2B-Think & BM25 & 13832 & 15728 & 1.14 & 167.9 & 201.9 \\
2B-Think & Embed & 12702 & 14958 & 1.18 & 113.4 & 144.6 \\
\midrule
4B-Instr & --- & 2815 & 3198 & 1.14 & 17.6 & 23.6 \\
4B-Instr & BM25 & 11970 & 13451 & 1.12 & 66.2 & 64.8 \\
4B-Instr & Embed & 10780 & 11966 & 1.11 & 51.4 & 79.5 \\
4B-Think & --- & 5119 & 6067 & 1.19 & 62.2 & 82.8 \\
4B-Think & BM25 & 10799 & 14011 & 1.30 & 166.1 & 250.8 \\
4B-Think & Embed & 9837 & 12700 & 1.29 & 124.6 & 187.5 \\
\midrule
8B-Instr & --- & 3475 & 4704 & 1.35 & 50.7 & 80.3 \\
8B-Instr & BM25 & 12602 & 16947 & 1.34 & 24.6 & 37.2 \\
8B-Instr & Embed & 10806 & 11351 & 1.05 & 19.7 & 23.1 \\
8B-Think & --- & 5019 & 6234 & 1.24 & 81.6 & 120.3 \\
8B-Think & BM25 & 13709 & 15907 & 1.16 & 247.1 & 344.9 \\
8B-Think & Embed & 11052 & 13890 & 1.26 & 176.9 & 260.7 \\
\bottomrule
\end{tabular}
\caption{Cost on full dataset (F) vs.\ rare entities (R, bottom-5\% popularity). R/F: token ratio.}
\label{tab:cost-rare}
\end{table*}

% \paragraph{Index construction.}
% The BM25 index is built from our 7.4M-article Wikipedia parquet index (243\,MB) using the \texttt{bm25s} library with English stemming, producing a 293\,MB sparse index. The embedding index uses \texttt{intfloat/multilingual-e5-large-instruct} (570M parameters) to encode all titles and descriptions into 1024-dimensional vectors stored in a FAISS \texttt{IndexFlatIP} index ($\sim$30\,GB at runtime). Both indices are constructed once and reused across all configurations.

% ===================================================
% APPENDIX K: SEARCH QUERY ANALYSIS
% ===================================================

\subsection{Search Query Analysis}
\label{app:query-content}

\paragraph{Query length.}
Table~\ref{tab:query-length} shows mean query length across configurations. All models produce queries of similar average length (14--21 characters), suggesting that query length is not a differentiating factor in the Think--Instruct gap.

\begin{table*}[t]
\centering
\small
\begin{tabular}{llrr}
\toprule
\textbf{Model} & \textbf{RAG} & \textbf{Mean Chars} & \textbf{Mean Words} \\
\midrule
2B-Instr & BM25 & 18.9 & 3.0 \\
2B-Instr & Embed & 14.6 & 2.4 \\
2B-Think & BM25 & 7.8 & 1.3 \\
2B-Think & Embed & 7.8 & 1.3 \\
\midrule
4B-Instr & BM25 & 16.2 & 2.4 \\
4B-Instr & Embed & 15.9 & 2.4 \\
4B-Think & BM25 & 10.3 & 1.6 \\
4B-Think & Embed & 9.5 & 1.5 \\
\midrule
8B-Instr & BM25 & 18.5 & 2.8 \\
8B-Instr & Embed & 17.5 & 2.6 \\
8B-Think & BM25 & 14.5 & 2.2 \\
8B-Think & Embed & 12.8 & 2.0 \\
\bottomrule
\end{tabular}
\caption{Mean search query length across configurations.}
\label{tab:query-length}
\end{table*}

\paragraph{Query transition patterns.}
Table~\ref{tab:query-patterns} categorizes each consecutive query pair by transition type. The dominant patterns are \emph{refinement} (adding disambiguation context to the previous query, 30--55\% of transitions) and \emph{variation} (moderate change, 25--50\%). Verbatim \emph{repetition} is rare in early searches but increases for Instruct models in long search chains (see Figure~\ref{fig:query-evolution}). \emph{Pivot} transitions (completely different query) occur 5--15\% of the time, typically when initial approaches fail.

\begin{table*}[t]
\centering
\small
\begin{tabular}{llrrrrr}
\toprule
\textbf{Model} & \textbf{RAG} & \textbf{Refine} & \textbf{Variation} & \textbf{Pivot} & \textbf{Repeat} & \textbf{Simplify} \\
\midrule
2B-Instr & BM25 & 21 & 12 & 18 & 46 & 3 \\
2B-Instr & Embed & 34 & 15 & 19 & 28 & 3 \\
2B-Think & BM25 & 2 & 1 & 96 & 0 & 0 \\
2B-Think & Embed & 6 & 0 & 81 & 12 & 0 \\
\midrule
4B-Instr & BM25 & 38 & 38 & 14 & 7 & 3 \\
4B-Instr & Embed & 46 & 37 & 14 & 2 & 2 \\
4B-Think & BM25 & 62 & 12 & 21 & 2 & 3 \\
4B-Think & Embed & 60 & 9 & 24 & 2 & 6 \\
\midrule
8B-Instr & BM25 & 35 & 41 & 19 & 3 & 3 \\
8B-Instr & Embed & 40 & 38 & 19 & 1 & 2 \\
8B-Think & BM25 & 56 & 19 & 13 & 9 & 3 \\
8B-Think & Embed & 58 & 17 & 16 & 6 & 3 \\
\bottomrule
\end{tabular}
\caption{Query transition patterns (\% of all consecutive query pairs). \emph{Refine}: previous query is substring of current (context added). \emph{Variation}: moderate change. \emph{Pivot}: completely different query (Jaccard $>$ 0.8). \emph{Repeat}: verbatim copy. \emph{Simplify}: current is substring of previous.}
\label{tab:query-patterns}
\end{table*}

\subsection{Second Model Family: GLM-4.6V-Flash}
\label{app:second-family}

All controlled factorial experiments in the main paper use Qwen3-VL. We evaluate a second family, GLM-4.6V-Flash~\cite{vteam2025glm45vglm41vthinkingversatilemultimodal}, as a cross-family check of whether retrieval also improves rare-entity accuracy in a reasoning configuration.

Without retrieval, the two families start at parity: 84.5 for GLM-4.6V-Flash thinking mode vs.\ 84.2 for Qwen3-VL-8B-Thinking, each under its vendor-recommended decoding configuration. On published multimodal-reasoning benchmarks the two families are comparable, each leading on some: MMMU~\cite{yue2024mmmumassivemultidisciplinemultimodal} 74.1 vs.\ 71.1, MathVista~\cite{lu2024mathvistaevaluatingmathematicalreasoning} 82.7 vs.\ 81.4. The families also differ in how reasoning is implemented. Qwen realizes reasoning as two separately post-trained checkpoints (Thinking and Instruct), while GLM exposes reasoning as an inference-time toggle on a single set of weights.

Table~\ref{tab:second-family} shows the effect of adding embedding retrieval for each configuration. For GLM's thinking mode, adding embedding retrieval improves all 15 rare-entity slices, with statistically significant gains on 11. This shows that the rare-entity retrieval benefit is not limited to Qwen. However, GLM's non-thinking mode could not sustain the retrieval loop, so this experiment does not test whether reasoning is necessary for those gains. The controlled Qwen factorial therefore provides the evidence for the reasoning-by-retrieval interaction. On the full set, where parametric knowledge already covers most entities, GLM with retrieval loses 2.7\% compared to its no-retrieval counterpart.

\begin{table*}[t]
\centering
\small
\begin{tabular}{lll}
\toprule
\textbf{Configuration} & \textbf{Full set (no retrieval $\rightarrow$ embed.)} & \textbf{Rare subsets (no retrieval $\rightarrow$ embed.)} \\
\midrule
Qwen3-VL-8B-Instruct & 83.5 $\rightarrow$ 82.9 ($-0.5$, n.s.) & 41.5--60.1 $\rightarrow$ 59.8--74.1 ($+8.2$ to +21.6) \\
Qwen3-VL-8B-Thinking & 84.2 $\rightarrow$ 87.9 (+3.8, $p<10^{-10}$) & 41.5--54.2 $\rightarrow$ 57.1--71.1 (+8.6 to +18.3) \\
GLM-4.6V-Flash, non-thinking & 82.7 $\rightarrow$ search loop not sustained\textsuperscript{1} & 46.9--62.4 $\rightarrow$ --- \\
GLM-4.6V-Flash, thinking & 84.5 $\rightarrow$ 81.8 ($-2.7$, $p<10^{-7}$) & 46.9--62.4 $\rightarrow$ 57.1--73.1 (+3.4 to +15.8) \\
\bottomrule
\end{tabular}
\caption{Accuracy (\%) with and without embedding retrieval, for Qwen3-VL and GLM-4.6V-Flash. Deltas are calculated from unrounded scores. Rare-entity ranges span 15 bottom-5\% slices. \textsuperscript{1}GLM entered the forced search tool-call format but degenerated into token repetition until exhausting the generation budget, on about 15\% of examples, at both the original and a doubled budget; we therefore report this configuration as unable to sustain the search loop rather than as an accuracy number.}

\label{tab:second-family}
\end{table*}

Qwen's reasoning checkpoint generates roughly 4.5$\times$ the completion tokens per example of GLM's thinking mode. This difference may help explain why Qwen converts retrieval into a full-set gain while GLM pays a modest cost from anchoring on retrieved candidates.

\subsection{Licenses for Artifacts Used and Released}
\label{app:licenses}

\paragraph{Models.} We use Qwen3-VL~\citep{qwen3technicalreport} under the Apache License 2.0, and the \texttt{multilingual-e5-large-instruct} embedding model~\citep{wang2024multilinguale5} under the MIT License. The CulturalPangea-7B baseline~\citep{nyandwi2025groundingmultilingualmultimodalllms} is used under the Apache License 2.0. The mGENRE baseline~\citep{decao2021multilingualautoregressiveentitylinking} is used under the MIT License, and the GEMEL baseline~\citep{shi-etal-2024-generative} is used under the terms of its original release.

\paragraph{Software.} We use FAISS~\citep{douze2024faiss} under the MIT License, BM25S~\citep{bm25s} under the MIT License, and SGLang~\citep{zheng2024sglangefficientexecutionstructured} under the Apache License 2.0.

\paragraph{Data.} We use the MERLIN benchmark~\citep{ramamoorthy2025merlintestbedmultilingualmultimodal} for evaluation in accordance with the terms set by its authors. English Wikipedia article content is used under the CC BY-SA 4.0 License, and Wikidata structural metadata is used under the CC0 1.0 Public Domain Dedication.

\paragraph{Released artifacts.} MERLIN-Rare consists of subsets of the existing MERLIN benchmark identified by structural-rarity metrics computed from Wikidata and is released under CC BY-SA 4.0. Our code is released under the MIT License.

All artifacts are used consistently with their intended research purposes.

\end{document}